\documentclass[preprint,3p]{elsarticle}

\usepackage{graphicx}
\usepackage{caption}
\usepackage{subcaption}
\usepackage{comment}
\usepackage{amssymb}
\usepackage{amsmath}
\usepackage{dirtytalk}
\usepackage{algorithm}
\usepackage{algorithmic}
\biboptions{numbers,sort&compress}
\usepackage{lineno}

\usepackage{xcolor}
\usepackage{hyperref}
\usepackage{siunitx}
\usepackage{tikz}
\usepackage{pgfplots}
\usetikzlibrary{matrix}
\usetikzlibrary{plotmarks}
\pgfplotsset{compat=newest}
\usepackage{makecell}
\usepackage[export]{adjustbox}

\newcommand{\mc}[1]{\mathcal{#1}}

\newcommand{\bs}[1]{\boldsymbol{#1}}

\usepackage{amsmath}

\DeclareMathOperator*{\argmin}{arg\,min}

\hypersetup{      
    colorlinks=true,                
    linkcolor= blue,                
    filecolor=black,
    citecolor=blue,
    linkbordercolor=blue
}
\graphicspath{ {./Figures/} }
\definecolor{brandeisblue}{rgb}{0.0, 0.44, 1.0}
\definecolor{brightpink}{rgb}{1.0, 0.0, 0.5}
\journal{ASCE}
\begin{document}

\setlength{\baselineskip}{14pt}

\begin{frontmatter}


\title{A Physics Informed Neural Network Approach to Solution and Identification of Biharmonic Equations of Elasticity}




\author[civilUNSW]{M. Vahab}
\author[civilMIT, civiUBC]{E. Haghighat}
\cortext[cor]{Corresponding author:}
\ead{ehsanh@mit.edu}
\author[civilSHARIF]{M. Khaleghi}
\author[civilUNSW]{N. Khalili}

\address[civilUNSW]{School of Civil and Environmental Engineering, The University of New South Wales, Sydney 2052, Australia}
\address[civilMIT]{Massachusetts Institute of Technology, Cambridge, MA, USA}
\address[civilUBC]{University of British Columbia, Vancouver, BC, Canada}
\address[civilSHARIF]{Department of Civil Engineering, Center of Excellence in Structures and Earthquake Engineering, Sharif University of Technology, Tehran, Iran }

\begin{abstract}
We explore an application of the Physics Informed Neural Networks (PINNs) in conjunction with Airy stress functions and Fourier series to find optimal solutions to a few reference biharmonic problems of elasticity and elastic plate theory. Biharmonic relations are fourth-order partial differential equations (PDEs) that are challenging to solve using classical numerical methods, and have not been addressed using PINNs. Our work highlights a novel application of classical analytical methods to guide the construction of efficient neural networks with the minimal number of parameters that are very accurate and fast to evaluate. In particular, we find that enriching feature space using Airy stress functions can significantly improve the accuracy of PINN solutions for biharmonic PDEs. 
\end{abstract}


\end{frontmatter}

\section{Introduction}
\label{S:1 (introduction)}

Deep learning (DL) has become a thriving framework that manifest in various fields including  speech/ handwriting recognition \cite{graves2013speech}, image processing/classification \cite{chan2015pcanet}, medical diagnoses \cite{razzak2018deep}, and fundamental scientific researches \cite{reichstein2019deep}. In engineering and science, DL has emerged as a promising alternative to emulate and capture the patterns, at primitive levels, as well as to predict complicated challenging interactive scenarios, at best, which is both evenly cumbersome and extremely time consuming to the cutting-edge simulators. DL is elaborated successfully in an increasing number of areas, including geo/material sciences \cite{pilania2013accelerating,bergen2019machine,shi2019deep}, solid/fluid/thermo mechanics \cite{brenner2019perspective,wu2018deep,zhu2021machine,im2021neural,anitescu2019artificial,bhatnagar2019prediction,yang2019conditional,minh2020surrogate,boso2020drug,lefik2009artificial}, and reservoir/electrical/chemical engineering \cite{zhang2018modeling,bouktif2018optimal,mayr2016deeptox,dehghani2021ann,lu2019data,raissi2019parametric,bahmani2021kd}, to name a few.

A common difficulty with the state-of-the-art machine learning (ML) techniques has been due to the extremely complex and expensive data acquisition in a vast majority of complex engineering / scientific systems \cite{deshpande2004model}. This threatens the feasibility and reliability of machine learning and renders conclusion and decision making a formidable task, if not impossible. A promising remedy in dealing with such problems is the prior physical knowledge, typically accessible in forms of Ordinary/Partial Differential Equations (i.e., ODEs/PDEs), which plays the role of a regularization agent to consolidate the admissibility of the solution, despite access to limited data. The first glimpse of the promising performance of prior information embedded deep learning in solution and discovery of high-dimensional PDEs has been due to the very recent contributions by Owhadi \cite{owhadi2015bayesian}, Han et al. \cite{han2018solving}, Bar-Sinai et al. \cite{bar2019learning}, Rudy et al. \cite{rudy2017data}, and Raissi et al. \cite{raissi2019physics}, which is now known as Physics-Informed Neural Networks (PINNs). PINNs are constructed and trained on the basis of a series of loss functions, endowed with the system of ODEs/PDEs and the corresponding Dirichlet / Neumann Boundary Conditions (BCs) or Initial Values (IVs), which govern the problem under consideration. PINNs are privileged in comparison to the preceding endeavors in the inclusion of the prior information as a result of the appropriate choice of network architecture, algorithmic advances (e.g., graph-based automated differentiation \cite{baydin2018automatic}), and software developments (e.g., TensorFlow \cite{abadi2016tensorflow}, Keras \cite{chollet2015keras}). Recently, PINNs has been successfully applied to the solution and discovery in fluid \cite{haghighat2021sciann,mao2020physics,sahli2020physics,kharazmi2021hp,mao2020physics}/solid \cite{haghighat2020deep,haghighat2020nonlocal, guo2020energy, zhang2021hierarchical,yang2019predictive}/pore \cite{yang2019adversarial,he2020physics,lou2020physics,rao2021physics}/thermo mechanics \cite{zhu2021machine,cai2021physics}, Eikonal equations \cite{waheed2021pinneik} (for a detailed review, see \cite{karniadakis2021physics}). 

In this study, we emphasize the novel elaboration of PINNs to the solution of a range of benchmark examples from the theory of elasticity and elastic plates, governed by the fourth-order Biharmonic equation. In the context of PINNs, the space of admissible solutions is approximated by general multi-layer neural networks, and the solution strategy is to minimize a loss function that includes the Mean-Squared Error norm (MSE) of the governing differential equations in conjunction with the complementary BCs/IVs, across randomly selected sampling points. The inherent capabilities of PINNs are explored with the discovery of parametric solutions based on the analytical closed-form solutions in the theory of elasticity. We examine the performance of PINNs in the solution and parametric study of elastic thin plates subjected to external loading. We find that, while setting up PINNs for different problems is relatively straightforward, training PINNs to reach a desired level of accuracy is an extremely hard problem. There are two reasons for such a poor performance; first, is associated with the use of first-order optimization methods; and second, with neural network architecture. However, we find that if we custom-design the network with \emph{features} leveraged from the classical analytical solutions of the Biharmonic equation, including Airy functions and Taylor series, we can improve their performance significantly. 

The paper is organized as follows: In section 2, we briefly describe the fundamentals of PINNs. In section 3, we elaborate PINNs in the solution of the Lamé problem and semi-infinite foundation, which are two well-known benchmark problems in the theory of elasticity. Section 4 is dedicated to the application of PINNs to the study of a selection of problems from the theory of elastic plates. Concluding remarks are presented in section 5. All examples presented in the current study are open-accessed and can be downloaded from \href{https://github.com/sciann/sciann-applications}{https://github.com/sciann/sciann-applications}.

\section{Physics-Informed Neural Networks}
\label{S:2 (PINN)}

Over the past few decades, abundant efforts has been conducted in relation to predictive physical modelling using machine learning approaches (e.g., support vector machines \cite{hearst1998support}, Gaussian processes \cite{rasmussen2003gaussian}, feed-forward \cite{svozil1997introduction}/convolutional \cite{gu2018recent}/recurrent neural networks \cite{medsker2001recurrent}). Most classical approaches employ deep learning as a black-box data-driven tool and demand a significant amount of data for training. This is commonly conceived as a big drawback for engineering applications due to restrictions with data accessibility, validity, noises, and other uncertainties. A beneficial remedy has been achieved thanks to the pioneering so-called physics-informed neural networks (PINNs), which embody the prior knowledge – typically in the forms of ODEs/PDEs – into the training process. By exploiting a feed-forward architecture in conjunction with Automatic Differentiation \cite{baydin2018automatic}, the physics-informed neural networks are constructed and trained to satisfy the underlying governing equations and, therefore, demand fewer data. 

\begin{figure}[!t]
\centering
\begin{minipage}{0.8\textwidth}
    \centering
    \includegraphics[width=0.9\linewidth]{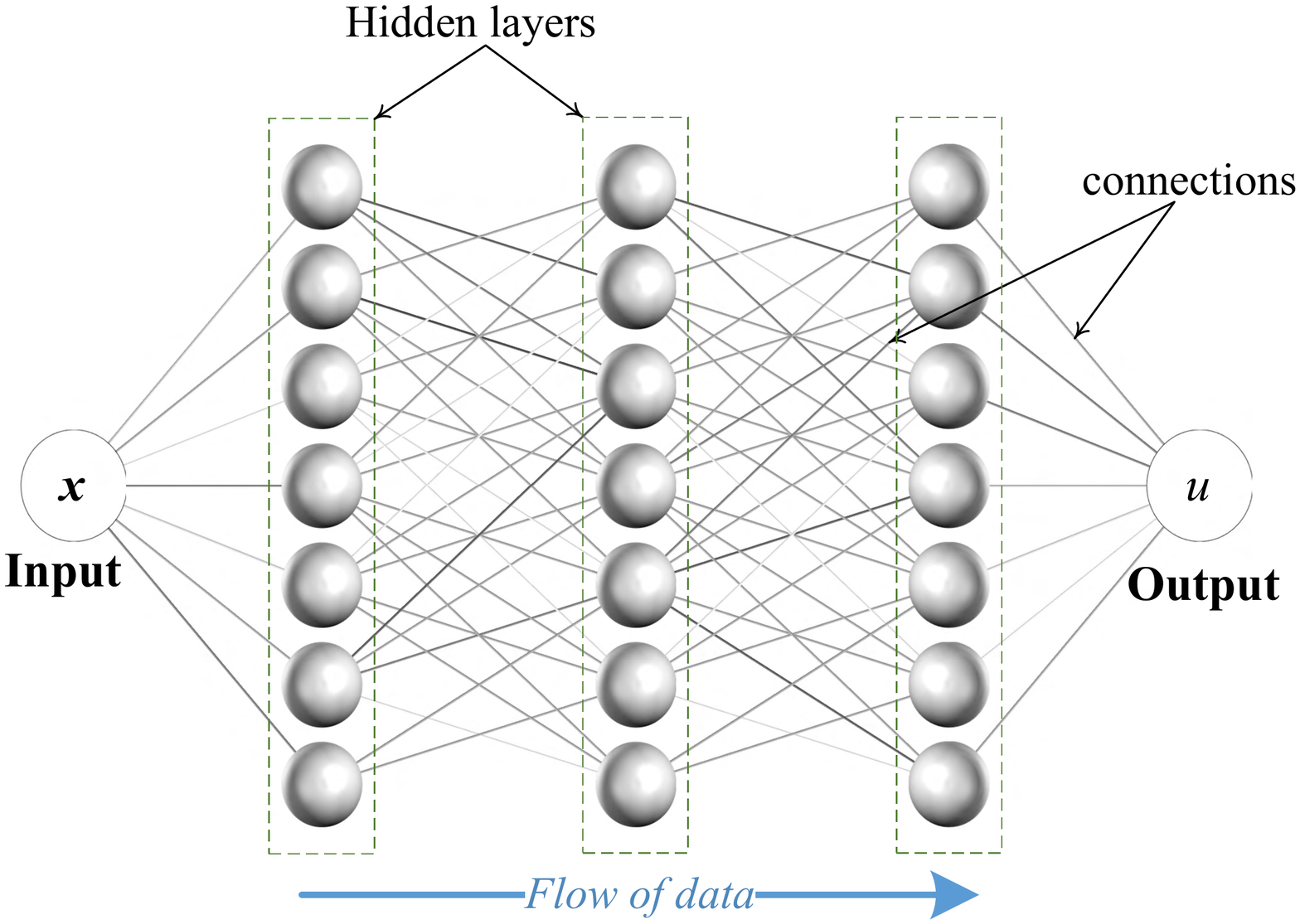}
    \end{minipage}
\caption{Standard PINN architecture, defining the mapping $u:\boldsymbol{x}\mapsto\mathcal{N}_u(\boldsymbol{x};\mathbf{W},b)$.}
   \label{fig:ann_architecture}
\end{figure}

 Let us consider a steady state partial differential operator $\mc{P}$ applied on a scalar solution variable $u$, i.e., $\mc{P}u(\bs{x}) = f(\bs{x})$, subjected to the boundary condition $\mc{B}u(\partial \bs{x}) = g(\partial\bs{x})$, with $\bs{x}\in\mathbb{R}^d$ and $d$ as the spatial dimension of the problem. According to the PINNs, the solution variable $u$ is approximated using a feed-forward neural network, with inputs $\bs{x}$ and outputs $u$. Consider a nonlinear map $\Sigma$ defined as $\Sigma^i(\hat{\bs{x}}^i) := \hat{\bs{y}}^i = \sigma^i(\bs{W}^i \cdot \hat{\bs{x}}^i + \bs{b}^i)$, with $\bs{W}, \bs{b}$ as weights and baiases of this transformation, and $\sigma$ as a nonlinear transformer. Such a nerual network is then expressed mathematically as 
\begin{equation}
\label{eq:NN}
u = \Sigma^L \circ \Sigma^{L-1} \circ \dots \circ \Sigma^0(\bs{x})
\end{equation}
where $\circ$ represents the compositional construction of the network and pictured graphically in Fig.\ref{fig:ann_architecture}. Such a network defines a mapping between input and output variables. $\bs{W}^i, \bs{b}^i$ form parameters of this transformation, equivalent to the degrees of freedom in a finite element discretization, and are all collected in $\bs{\theta} = \bigcup_{i=0}^L (\bs{W}^i, \bs{b}^i)$. For the internal layers, hyperbolic-tangent is often used as the nonlinear transformer $\sigma^i$, a.k.a. \emph{activation function}, and for the last layer, it is often kept \emph{linear} for regression-type problems, which we account here.

According to the PINNs, the parameters of this neural network are identified by minimizing a multi-objective loss function, consisted of the error associated with PDE residual and boundary conditions, and it is constructed as 
\begin{equation}
\label{eq:loss}
\mc{L}(\bs{x}; \bs{\theta}) = \sum \lambda_i \mc{L}_i = \lambda_1 \left\| \mc{P}u - 0 \right\| _{\Omega} + \lambda_2\left\| \mc{B}u - g \right\|_{\partial \Omega} + \dots 
\end{equation}
where $\mc{L}$ is a loss function. Since we deal with regression type problems, the mean squared error norm is used as the loss function, i.e., $\left\| \circ \right\| = \text{MSE}(\circ)$. Note $\lambda_i$'s are weights associated with each loss term. The optimal values of the networks parameters, i.e., $\bs{\theta}^*$, are then identified by optimizing this loss function as 
\begin{equation}
\label{eq:loss_sum}
\bs{\theta}^* = \argmin_{\bs{\theta}\in\mathbb{R}^D} \mc{L}(\bs{X}; \bs{\theta})
\end{equation}
with $D$ as the total number of trainable parameters, and $\bs{X}\in\mathbb{R}^{N\times d}$ is the set of $N$ collocation points used to optimize this loss function. 

As we find, to evaluate the loss function \eqref{eq:loss}, we need to evaluate derivatives of the $u$ at different training points. Instead of numerical differentiation used in earlier works, PINNs leverage automatic differentiation that is readily available in modern deep learning frameworks \cite{chollet2015keras}. This enables the explicit inclusion of any partial differential operator $\mc{P}$ or Neumann boundary conditions acting on the field variable $u$.

In this study, we use the open-source python API SciANN \cite{haghighat2021sciann}, to set up and train the PINN architectures. SciANN is optimized to perform scientific machine learning with minimal effort. Implemented on the reputed deep-learning packages Tensorow \cite{abadi2016tensorflow} and Keras \cite{chollet2015keras}, it inherits their functionalities in terms of graph-based automatic differentiation and high-performance computing. For the problems showcased in the following sections, unless stated otherwise, training is carried out using the Adam optimization scheme \cite{kingma2014adam}. The network hyperparameters such as the number of nodes per layer or the number of hidden layers are described separately for each problem.


\section{PINN Solution of Elasticity Problems}
\label{S:5 (Application of PINNs to Theory of Elasticity)}

In this section, PINNs are employed to the solution of problems in the theory of elasticity. The partial differential equation of the Airy stress function, as a robust solution procedure in the theory of elasticity, is briefly explained and is applied for the solution of two benchmark examples in the field.

\subsection{Theory of Elasticity}
\label{S:2-1 (SciANN)}

In the theory of elasticity, the solution field can be obtained by utilizing a variety of stress functions. The Airy stress function \cite{Airy} is a simplified form of the more general Maxwell stress function \cite{maxwell1890scientific}, which was originally introduced for the solution of two-dimensional elasticity problems. The idea was to develop a series of scalar representative stress functions, that simultaneously satisfy the equilibrium equation and compatibility requirements \cite{timoshenko412theory,boresi2010elasticity}. Neglecting the body forces, for the sake of simplicity in presentation, it follows that \cite{katebi2010axisymmetric}

\begin{equation}
\label{eq:Equilibrium}
\nabla ^4\phi  = \left( {\frac{{{\partial ^2}}}{{\partial {r^2}}} + \frac{1}{r}\frac{\partial }{{\partial r}} + \frac{1}{{{r^2}}}\frac{{{\partial ^2}}}{{\partial {\theta ^2}}}} \right)\left( {\frac{{{\partial ^2}}}{{\partial {r^2}}} + \frac{1}{r}\frac{\partial }{{\partial r}} + \frac{1}{{{r^2}}}\frac{{{\partial ^2}}}{{\partial {\theta ^2}}}} \right)\phi  = 0,
\end{equation}
in which $\phi$ is the Airy stress function in polar coordinate system, i.e., $(r,\theta)$. Alternatively, this equation can be expressed in the Cartesian coordinate system as 
\begin{equation}
\label{eq:Phi-XY}
\nabla ^4\phi  = \left( {\frac{{{\partial ^2}}}{{\partial {x^2}}} +{\frac{\partial ^2}{\partial y^2}}} \right)\left( {\frac{{{\partial ^2}}}{{\partial {x^2}}} +{\frac{\partial ^2}{\partial y^2}}} \right)\phi = 0,
\end{equation}
where $(x,y)=r(\text{sin}\theta,\text{cos}\theta)$. The above relation is called biharmonic equation which describes the in-plane elasticity problems with a unified expression. Eq. \eqref{eq:Equilibrium} can be solved in conjunction with adequate boundary conditions over $\partial\Omega$. Using the Airy stress function, the stress components can be determined as

\begin{equation}
\begin{split}
\label{eq:NN}
{\sigma _r} &= \frac{1}{r}\frac{{\partial \phi }}{{\partial r}} + \frac{1}{{{r^2}}}\frac{{{\partial ^2}\phi }}{{\partial {\theta ^2}}},\\
{\sigma _\theta } &= \frac{{{\partial ^2}\phi }}{{\partial {r^2}}},\\
{\sigma _{r\theta }} &=  - \frac{\partial }{{\partial r}}\left( {\frac{1}{r}\frac{{\partial \phi }}{{\partial \theta }}} \right).
\end{split}
\end{equation}
Using the Hooke's law, the strain components are expressed as

\begin{equation}
\begin{split}
\label{eq:NN}
{\epsilon _r}&=\frac{1}{2\mu}\left(\sigma _{r}-\frac{1}{4} (3-\kappa)(\sigma _{r}+\sigma _{\theta})\right),\\
{\epsilon _\theta}&=\frac{1}{2\mu}\left(\sigma _{\theta}-\frac{1}{4} (3-\kappa)(\sigma _{r}+\sigma _{\theta})\right),\\
{\epsilon _{r\theta}}&=\frac{1}{2\mu}\sigma _{r\theta},
\end{split}
\end{equation}
where $\kappa$ is $3-4\nu$ or $(3-\nu)/(1+\nu)$ for plane strain or plane stress state, respectively. In the above relations, $\mu$ is Lamé' second parameter (identical to the shear modulus, i.e., $G$), and $\nu$ is Poisson's ratio.

\subsection{Lamé Problem}
\label{S:2-1 (SciANN)}

\begin{figure}[!t]
\centering
\begin{minipage}{0.7\textwidth}
    \centering
    \includegraphics[width=0.8\linewidth]{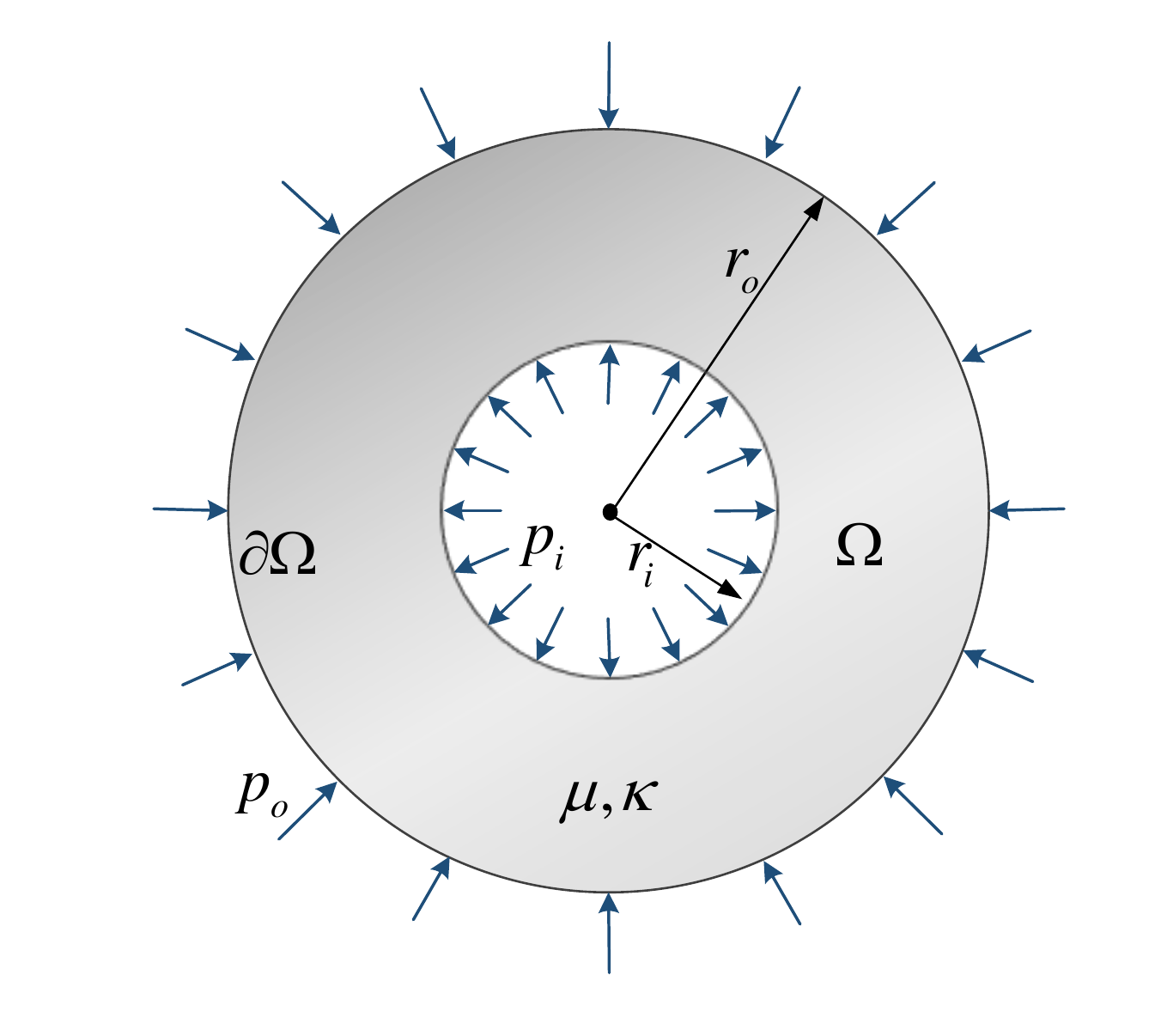}
    \end{minipage}
\caption{Lamé problem setup, subjected to internal and external boundary conditions $p_i$ and $p_o$, respectively.}
    \label{fig:definition-Lamé}
\end{figure}

In this example, the application of PINNs to the solution of circular annulus subjected to uniform external/internal pressure is illustrated (see Fig. \ref{fig:definition-Lamé}), which is known as Lamé problem \cite{lame1852lessons}. Considering the axisymmetry of the problem, the solution is independent of $\theta$. This, in turn, leads to a simplified form of the biharmonic equation (i.e., Eq. \eqref{eq:Equilibrium}) as

\begin{equation}
\label{eq:Phi-R}
\nabla ^4\phi  = \left( {\frac{{{\partial ^2}}}{{\partial {r^2}}} + \frac{1}{r}\frac{\partial }{{\partial r}} 
} \right)\left( {\frac{{{\partial ^2}}}{{\partial {r^2}}} + \frac{1}{r}\frac{\partial }{{\partial r}} } \right)\phi  = 0.
\end{equation}
The problem definition is accomplished by the introduction of the boundary conditions as $\sigma _r(r=r_i,\theta)=p_i$ and $\sigma _r(r=r_o,\theta) =p_o$, with  $p_i$ and $p_o$ being the pressure exerted towards the internal and external faces of the annulus, respectively.

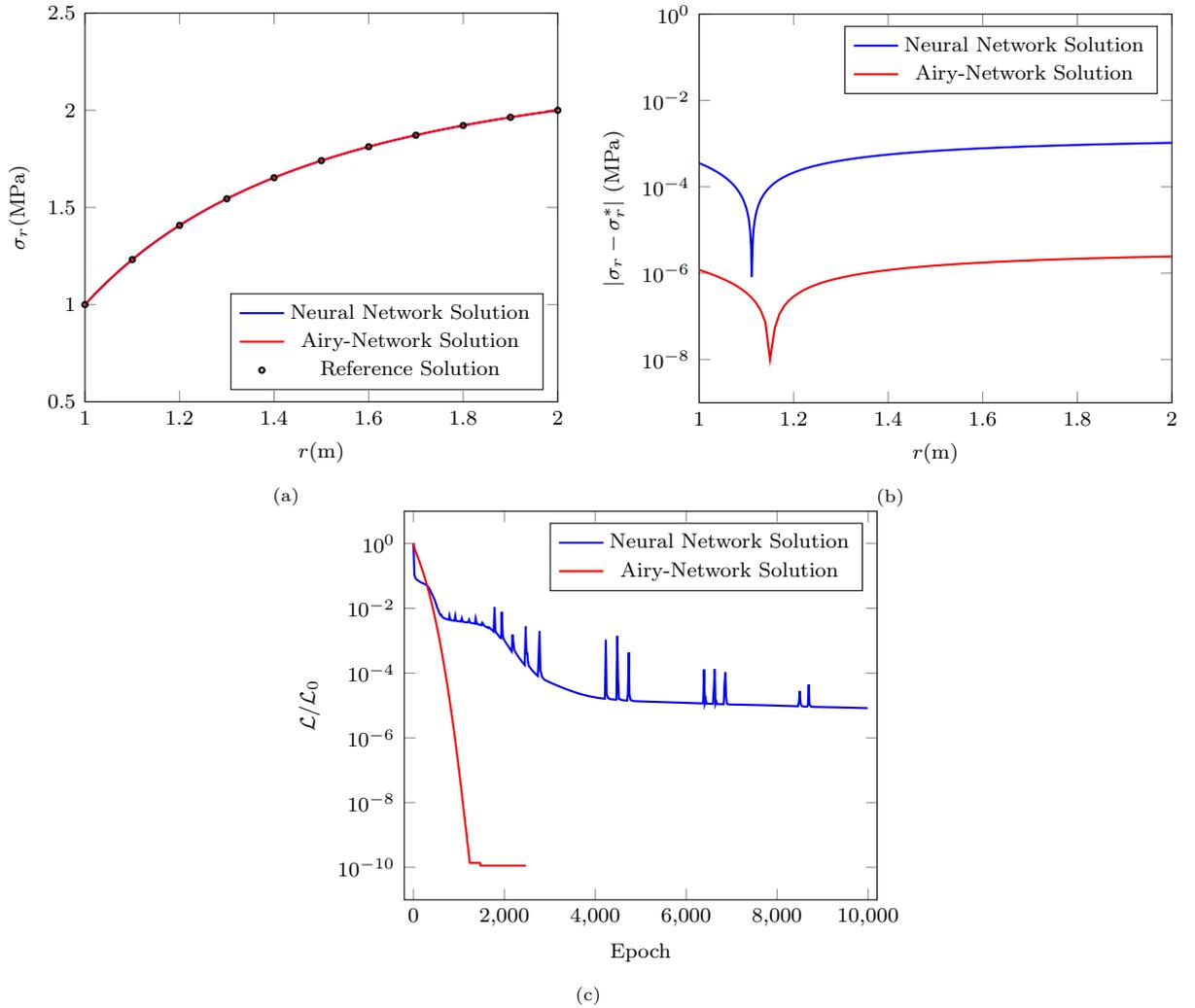
\begin{figure}[!b]
\centering
\begin{minipage}{0.49\textwidth}
    \centering
    \begin{tikzpicture}
    \begin{axis}[
        every axis plot/.append style={thick},
        xlabel={$r$(m)},
        ylabel={$\sigma_r$(MPa)},
        xmin=1, xmax=2,
        ymin=0.5, ymax=2.5,
        xtick={1.0,1.2,1.4,1.6,1.8,2.0},
        ytick={0.5,1.0,1.5,2.0,2.5},
        legend pos=south east,
        /tikz/font=\footnotesize,
        legend style={font=\footnotesize},
        width = 0.99\linewidth
    ]
    \addplot[color=blue,]
        table {./Data/Example1_Sr_R.dat};
        \addlegendentry{Neural Network Solution}
    \addplot[color=red,]
        table {./Data/Example1b_Sr.dat};
        \addlegendentry{Airy-Network Solution}
    \addplot[color=black, only marks, mark=o,mark repeat=10,mark size=1.0pt]
        table {./Data/Example1b_Sr_exact.dat};
        \addlegendentry{Reference Solution};
    
    \end{axis}
    \end{tikzpicture}
    \subcaption{}
    \label{fig:Lamé solution-a}
\end{minipage}
\begin{minipage}{0.49\textwidth}
    \centering
    \begin{tikzpicture}
    \begin{axis}[
        every axis plot/.append      style={thick},
        ylabel={ $\left|\sigma_r-\sigma_r^*\right|$ (MPa) },
        xlabel={$r$(m)},
        ymode=log,
        y tick label style={/pgf/number format/sci},
        xmin=1, xmax=2,
        ymin=0.000000001, ymax=1,
        scaled x ticks = false,
        xtick={1.0,1.2,1.4,1.6,1.8,2.0},
        ytick={0.0000000001,0.00000001,0.000001,0.0001,0.01,1},
        legend pos=north east,
        /tikz/font=\footnotesize,
        legend style={font=\footnotesize},
        width = 0.99\linewidth
    ]
    \addplot[color=blue]
        table {./Data/Example1_Sr-error.dat};
        \addlegendentry{Neural Network Solution};
    \addplot[color=red]
        table {./Data/Example1b_Sr-error.dat};
        \addlegendentry{Airy-Network Solution};
    \end{axis}
    \end{tikzpicture}
    \subcaption{}
    \label{fig:Lamé solution-b}
\end{minipage}
\begin{minipage}{0.49\textwidth}
    \centering
    \begin{tikzpicture}
    \begin{axis}[
        every axis plot/.append      style={thick},
        ylabel={$\mathcal{L}/\mathcal{L_\text{0}}$},
        xlabel={Epoch},
        ymode=log,
        y tick label style={/pgf/number format/sci},
        xmin=-200, xmax=10200,
        ymin=0.00000000001, ymax=10.0,
        scaled x ticks = false,
        xtick={0,2*10^3,4000,6000,8000,10000},
        ytick={0.0000000001,0.00000001,0.000001,0.0001,0.01,1},
        legend pos=north east,
        /tikz/font=\footnotesize,
        legend style={font=\footnotesize},
        width = 0.99\linewidth
    ]
    \addplot[color=blue]
        table {./Data/Example1_history.dat};
        \addlegendentry{Neural Network Solution};
    \addplot[color=red]
        table {./Data/Example1b_history.dat};
        \addlegendentry{Airy-Network Solution};
    \end{axis}
    \end{tikzpicture}
    \subcaption{}
    \label{fig:Lamé solution-c}
\end{minipage}
\caption{Deep learning solution for Lamé problem; a) profile of radial stresses $\sigma_r$, b) error norm of radial stresses, c) network training history.}
\label{fig:Lamé solution}
\end{figure}
In the theory of elasticity, it is shown that the solution to biharmonic Eq. \eqref{eq:Phi-R} on the basis of the above-mentioned boundary conditions is multi-valued. In order to circumvent this difficulty, the so-called consistency constraints of the displacement field need to be incorporated. A beneficial constraint is imposed through the definition of strains versus displacement field as \cite{timoshenko412theory}
\begin{equation}
\begin{split}
\label{eq:NN}
{\epsilon _r}&=\frac{\partial u_r}{\partial r},\\
{\epsilon _\theta}&=\frac{u_r}{r}+\frac{\partial u_\theta}{r \partial \theta}.\\
\end{split}
\end{equation}
Due to the axisymmetry of the problem, the tangential displacement $u_\theta$ must yield the identical values independent of $\theta$, or simply vanish (i.e., $u_\theta \equiv 0$).

Here, the solution to Lamé problem is investigated for $p_i=1 \text{ MPa}$ and $p_o=2 \text{ MPa}$ by means of the PINNs. Suppose the domain is delimited to $r_i=1 \text{ m}$ and $r_o=2 \text{ m}$, with material properties set to $G=100 \text{ GPa}$ and $ \nu=0.25$. To this end, it is required to define the unknown solutions using the below neural networks

\begin{equation}\label{eq:NN}
\begin{split}
\phi(r) &\simeq \mathcal{N}_{\phi}(r), \\
u_r &\simeq \mathcal{N}_{u_r}(r). \\
\end{split}
\end{equation}
Notably, the inclusion of the displacement field in the neural network architecture facilitates the imposition of the boundary conditions related to the consistency constraint in the strain field. The physics-informed loss terms of the total cost function are defined as

\begin{equation}
\begin{split}
\label{eq:loss_sum}
\mathcal{L}_\Omega &=\left\| \left( {\frac{{{\partial ^2}}}{{\partial {r^2}}} + \frac{1}{r}\frac{\partial }{{\partial r}} 
} \right)\left( {\frac{{{\partial ^2}}}{{\partial {r^2}}} + \frac{1}{r}\frac{\partial }{{\partial r}} } \right)\phi \right\|,\\
\mathcal{L}_\epsilon&=\left\|\frac{1}{2\mu}\left(r\sigma _{\theta}-\frac{1}{4} (3-\kappa)(\sigma _{r}+\sigma _{\theta})\right)-u_r\right\|,\\
\mathcal{L}_{\partial\Omega_1}&=\lambda_1 \left\|(\sigma_r-p_i)\left(r=r_\text{min}\right)\right\|,\\
\mathcal{L}_{\partial\Omega_2}&=\lambda_2 \left\|(\sigma_r-p_o)\left(r=r_\text{max}\right)\right\|.
\end{split}
\end{equation}
Training is performed over 2000 sampling points (same as batch size)  through 10000 epochs. It is noteworthy that in the above relation $\lambda_i$'s are mini-batch optimization parameters, which are set to relatively large values to guarantee consistent optimization on the boundaries.

As per the approximation functions for the unknown variables $\phi$ and $u_r$, two approaches are taken. In the first approach, they are approximated using two neural networks, which consists of 5 hidden layers and 20 neurons per layer, with hyperbolic-tangent utilized as the activation function. In the second approach, the Airy stress functions are adopted to construct the optimal approximation space. Therefore, the approximation function takes the form of 
\begin{equation}
\label{eq:Phi2}
\begin{split}
\phi(r)&=a_1\text{log}r+a_2 r^2\text{log}r+a_3r^2+a_4, \\
u_r(r)&=\frac{1}{E}(-\frac{a_1}{r}+2a_2r\text{log}r +2a_3r )-\frac{\nu}{E}\phi,
\end{split}
\end{equation}
in which $a_i$'s are the only parameters of the optimization problem, identified by minimizing the same physics-informed loss function. 

The results are presented in Fig. \ref{fig:Lamé solution}, where $\lambda_1$ and $\lambda_2$ are preset to large numbers $\lambda_1=\lambda_2=100$ so as to accurately enforce the boundary conditions. In this regard, the profiles of the radial stress component $\sigma_r$ and its absolute error with respect to the exact values (i.e., $\sigma_r^*$) are shown for the PINNs solution and compared against the analytical solution in the theory of elasticity \cite{timoshenko412theory}. Excellent agreement is observed between the present model and the reference solution, which demonstrates the capabilities of the PINNs in the solution of higher-order ODEs. In Fig. \ref{fig:Lamé solution-c}, the evolution of normalized loss versus epochs is depicted, which indicates the rapid convergence in the training of PINNs for the Lamé problem. It is evident from the solutions that the Airy-network results in superior convergence and accuracy characteristics, and therefore is the optimal choice for solving the Lamé problem. It should also be noted that due to the small number of parameters of the Airy-network, the optimization time per epoch is much faster than the case of the neural network, which is summed up to at least 10 times greater efficiency in total training duration.


\begin{figure}[!t]
\centering
\begin{subfigure}{0.49\textwidth}
    \centering
    \includegraphics[trim={0 5 0 5},width=0.95\linewidth]{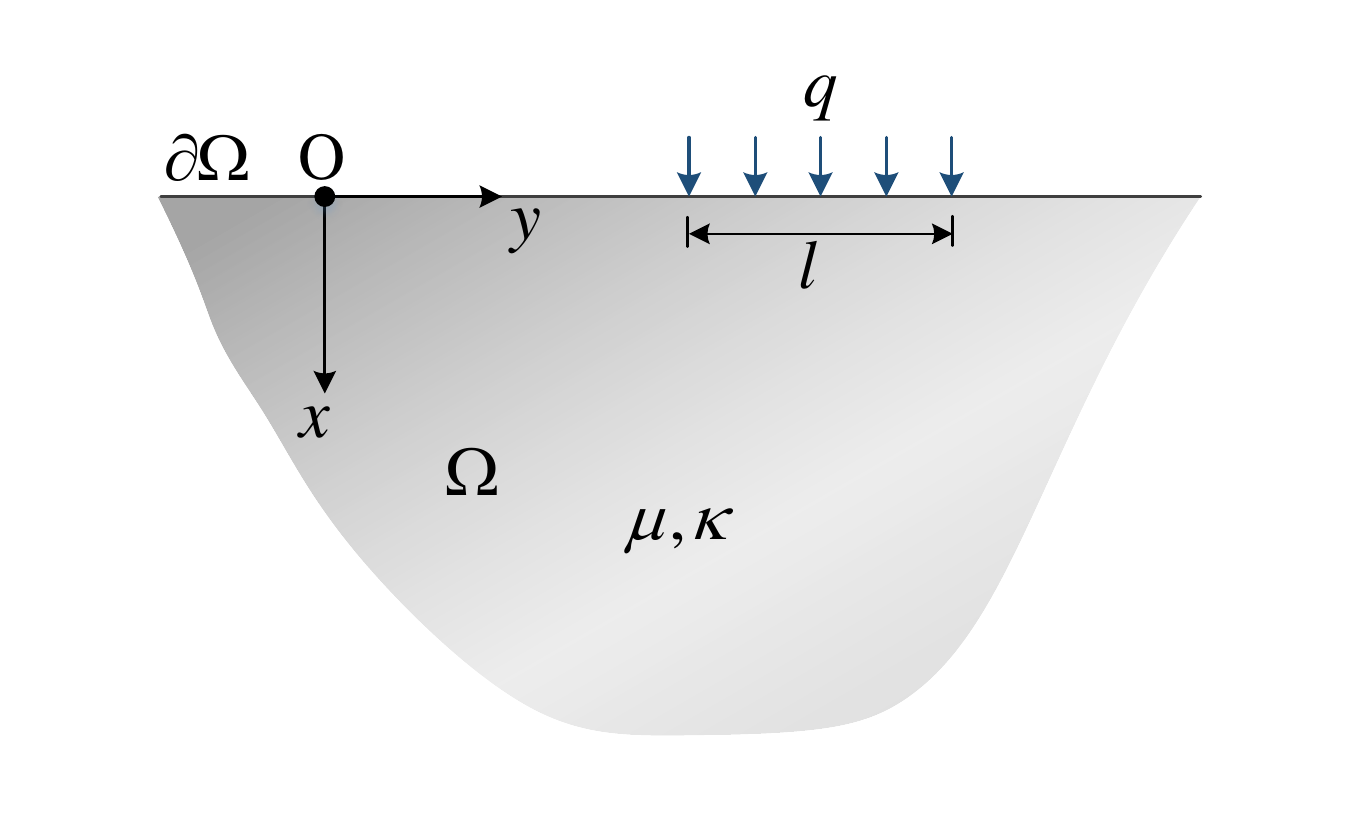}
    \label{fig:EX1-a}
    \caption{Finite pressure}
\end{subfigure}
\begin{subfigure}{0.45\textwidth}
    \centering
    \includegraphics[trim={0 0 0 0},width=0.95\linewidth]{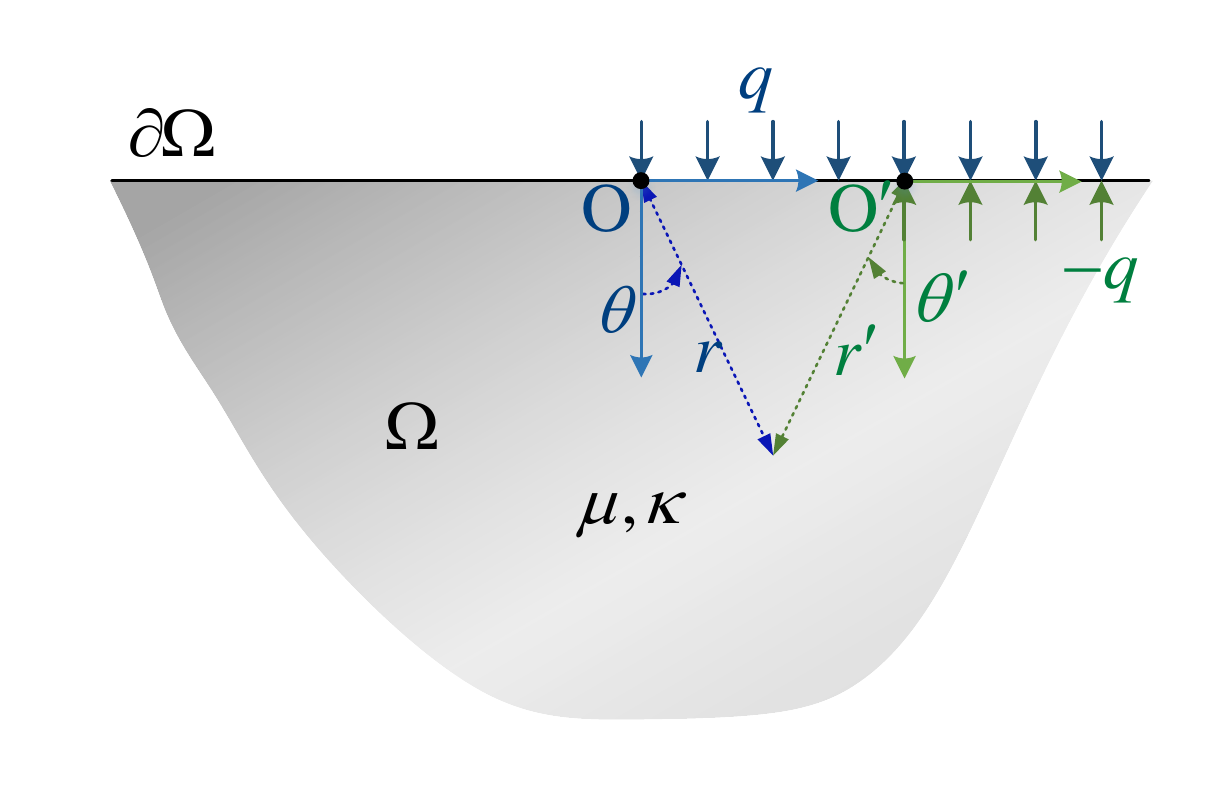}
    \caption{Infinite pressure}
    \label{fig:EX1-b}
\end{subfigure}
\caption{Problem definition for half-space subject to strip footing; a) Geometry and boundary conditions, b) Intermediate solutions.}
\label{fig:Foundation}
\end{figure}

\subsection{Foundation}
\label{S:3-3 (Footing)}

As the second case, we investigate the solution of a half-space subjected to uniform pressure over a finite interval of its boundary using PINNs, which is representative of a strip foundation. In the theory of elasticity, this problem is solved through the superposition of two half-spaces subjected to shifted semi-infinite uniform pressure along one-half of its boundary \textendash in reverse directions \textendash as shown in Fig. \ref{fig:Foundation}. Accordingly, the Airy stress function for the solution to half-space subjected to uniform pressure $q$ is defined as $\phi=qr^2 f(\theta)$ \cite{timoshenko412theory}. Inserting this definition into the biharmonic Eq. \eqref{eq:Equilibrium} yields

\begin{equation}
\label{eq:Equilibrium2}
\nabla ^4\phi  = \frac{1}{r^2} \frac{{{\partial ^2}}}{{\partial {\theta ^2}}} \left(
\frac{{{\partial ^2f(\theta)}}}{{\partial {\theta^2}}}+4f(\theta)
\right)=0.
\end{equation}

Here, the PINNs is applied to obtain the solution for $q=1 \text{ kPa}$ using the Airy-network architecture ${\phi(r,\theta) \simeq qr^2 \mathcal{N}_{\theta}(\theta)}$. Alternatively, the physics-informed solution to the general form of the biharmonic equation in the Cartesian coordinate system (i.e., Eq. \eqref{eq:Phi-XY})  is elaborated to construct a more general neural network to solve this problem. It is noteworthy that the description of the biharmonic Equation in Cartesian coordinates is preferred over the polar coordinates description (i.e., Eq. \eqref{eq:Equilibrium}) to avoid the problematic singularity of the equations with respect to $r$ at the origin. As a remedy in this example, the order of the biharmonic Eq. \eqref{eq:Phi-XY} is reduced through the change of variables as
\begin{equation}
\begin{split}
\label{eq:Phi-XY-red}
\nabla ^2\psi  &= \left( {\frac{{{\partial ^2}}}{{\partial {x^2}}} +{\frac{\partial ^2}{\partial y^2}}} \right)\psi = 0,\\
\psi  &= \left( {\frac{{{\partial ^2}}}{{\partial {x^2}}} +{\frac{\partial ^2}{\partial y^2}}} \right)\phi.
\end{split}
\end{equation}
In this fashion, two conjugate neural networks, i.e., ${\phi(x,y) \simeq \mathcal{N}_{\phi}(x,y)}$ and ${\psi(x,y) \simeq \mathcal{N}_{\psi}(x,y)}$, are employed to solve the biharmonic equation. For the sake of comparison a solo neural network, i.e., ${\phi(x,y) \simeq \mathcal{N}_{\phi}(x,y)}$, is also exercised by using the fourth-order form of the biharmonic equation. In all cases, 6 hidden layers with 20 neurons per layer are supposed for the construction of the neural networks, where the loss terms of the total cost function are described as
\begin{equation}
\begin{split}
\label{eq:loss_sum}
\mathcal{L}_\Omega&=\left\|  \nabla ^4\phi \right\|,\\
\mathcal{L}_{\partial\Omega_1}&=\lambda_1 \left\|(\sigma_\theta-q)\left(\theta=\theta_\text{max}\right)\right\|,\\
\mathcal{L}_{\partial\Omega_1}&=\lambda_2 \left\|\sigma_\theta\left(\theta=\theta_\text{min}\right)\right\|,\\
\mathcal{L}_{\partial\Omega_1}&=\lambda_3 \left\|\sigma_{r\theta}\left(\theta=\theta_\text{max}\right)\right\|,\\
\mathcal{L}_{\partial\Omega_1}&=\lambda_4 \left\|\sigma_{r\theta}\left(\theta=\theta_\text{min}\right)\right\|.\\
\end{split}
\end{equation}

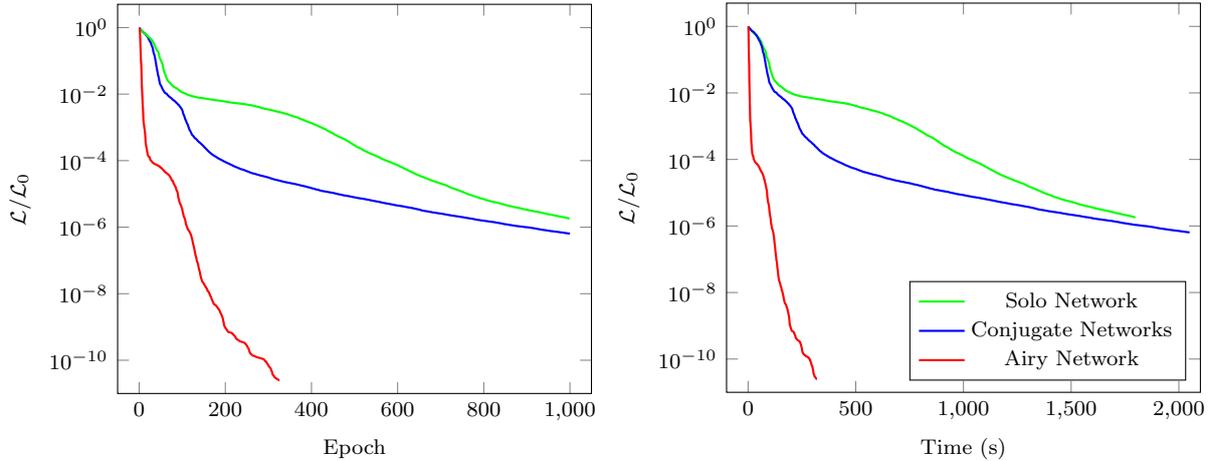
\begin{figure}[!t]
\centering
\begin{minipage}{0.48\textwidth}
    \begin{tikzpicture}
    \begin{axis}[
        every axis plot/.append      style={thick},
        ylabel={$\mathcal{L}/\mathcal{L_\text{0}}$},
        xlabel={Epoch},
        ymode=log,
        xmin=-50, xmax=1050,
        ymin=0.00000000001, ymax=5.0,
        xtick={0,200,400,600,800,1000},
        ytick={0.0000000001,0.00000001,0.000001,0.0001,0.01,1},
        legend pos=north east,
        /tikz/font=\footnotesize,
        legend style={font=\footnotesize},
        scaled x ticks = false,
        width = 0.99\linewidth,
        y tick label style={/pgf/number format/sci}
    ]
    \addplot[color=blue]
        table {./Data/Example2_history_NN.dat};
    \addplot[color=green]
        table {./Data/Example2_history_NN_4.dat};
     \addplot[color=red]
        table {./Data/Example2_history.dat};
    \end{axis}
    
    \end{tikzpicture}
\end{minipage}
\begin{minipage}{0.48\textwidth}
    \begin{tikzpicture}
    \begin{axis}[
        every axis plot/.append      style={thick},
        ylabel={$\mathcal{L}/\mathcal{L_\text{0}}$},
        xlabel={Time (s)},
        ymode=log,
        xmin=-100, xmax=2100,
        ymin=0.00000000001, ymax=5.0,
        xtick={0,500,1000,1500,2000},
        ytick={0.0000000001,0.00000001,0.000001,0.0001,0.01,1},
        legend pos=south east,
        /tikz/font=\footnotesize,
        legend style={font=\footnotesize},
        scaled x ticks = false,
        width = 0.99\linewidth,
        y tick label style={/pgf/number format/sci}
    ]
    \addplot[color=green]
        table {./Data/Example2_history_time_NN_4.dat};
        \addlegendentry{Solo Network};
    \addplot[color=blue]
        table {./Data/Example2_history_time_NN.dat};
        \addlegendentry{Conjugate Networks};
     \addplot[color=red]
        table {./Data/Example2_history_time.dat};
        \addlegendentry{Airy Network };
    \end{axis}
    \end{tikzpicture}
\end{minipage}
\caption{The network training history for semi-infinite foundation subjected to uniform loading along one-half of its boundary.}
\label{fig:Foundation_loss}
\end{figure}

Training is performed over a 100 by 100 grid on a half-space of radius $r_o=5 \text{ m}$ using full-batch decent. In this example, the SciPy (Scientific Python) BFGS (Broyden–Fletcher–Goldfarb–Shanno \cite{virtanen2020scipy}) optimization algorithm is exclusively employed for the training of the model. In Fig. \ref{fig:Foundation_loss}, the evolution of normalized loss versus epochs and time is presented. As can be seen, the training time for the Airy-network solution is significantly faster than the solo/conjugate neural networks. This is attributed to the beneficial use of the suggested analytical form that involves simpler architecture. In addition, it can be seen that the Airy-network solution has reached much greater accuracy for the fewer number of epochs since it is better aligned with the exact solution. Meanwhile, the conjugate neural networks feature a greater convergence rate (versus epochs) with respect to the solo neural network, yet at the expense of slightly longer training time.

\begin{figure}[!t]
\centering
\begin{subfigure}{0.40\textwidth}
    \centering
    \includegraphics[trim={0 0 0cm 0},width=0.99\linewidth]{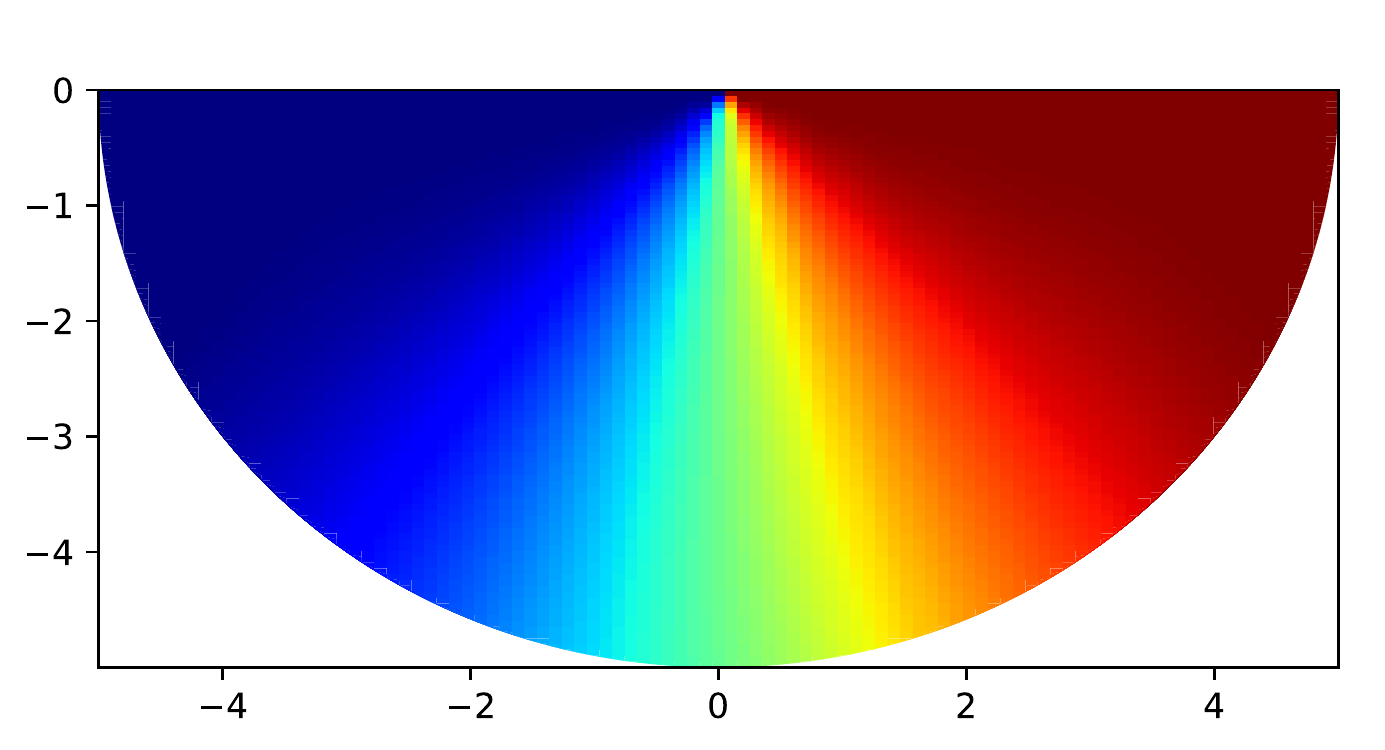}
    \caption{Conjugate networks solution}
    \label{fig:EX1-a}
\end{subfigure}
\begin{subfigure}{0.4\textwidth}
    \centering
    \includegraphics[trim={0 0 0cm 0},width=0.99\linewidth]{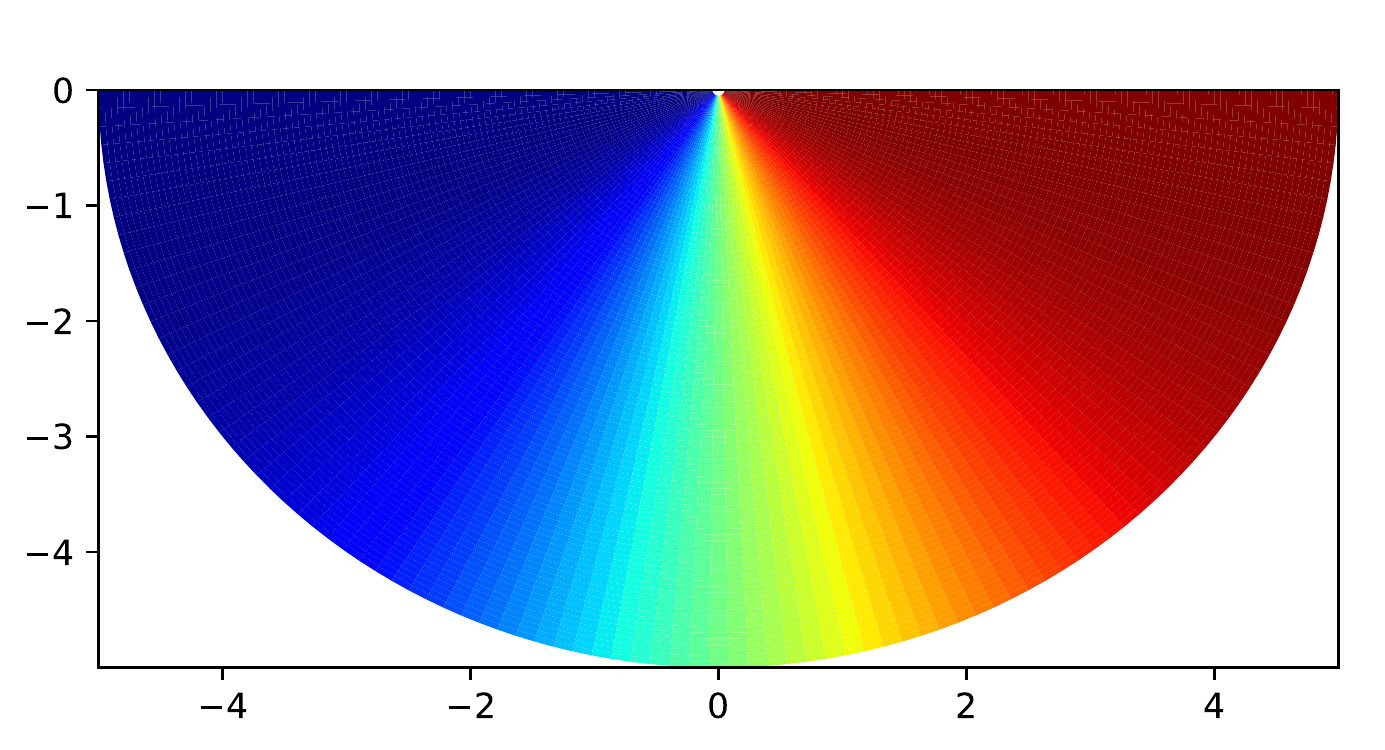}
    \caption{Airy-network solution}
    \label{fig:EX1-a}
\end{subfigure}
\begin{subfigure}{0.57\textwidth}
    \centering
    \includegraphics[trim={0 0 0cm 0},width=0.99\linewidth]{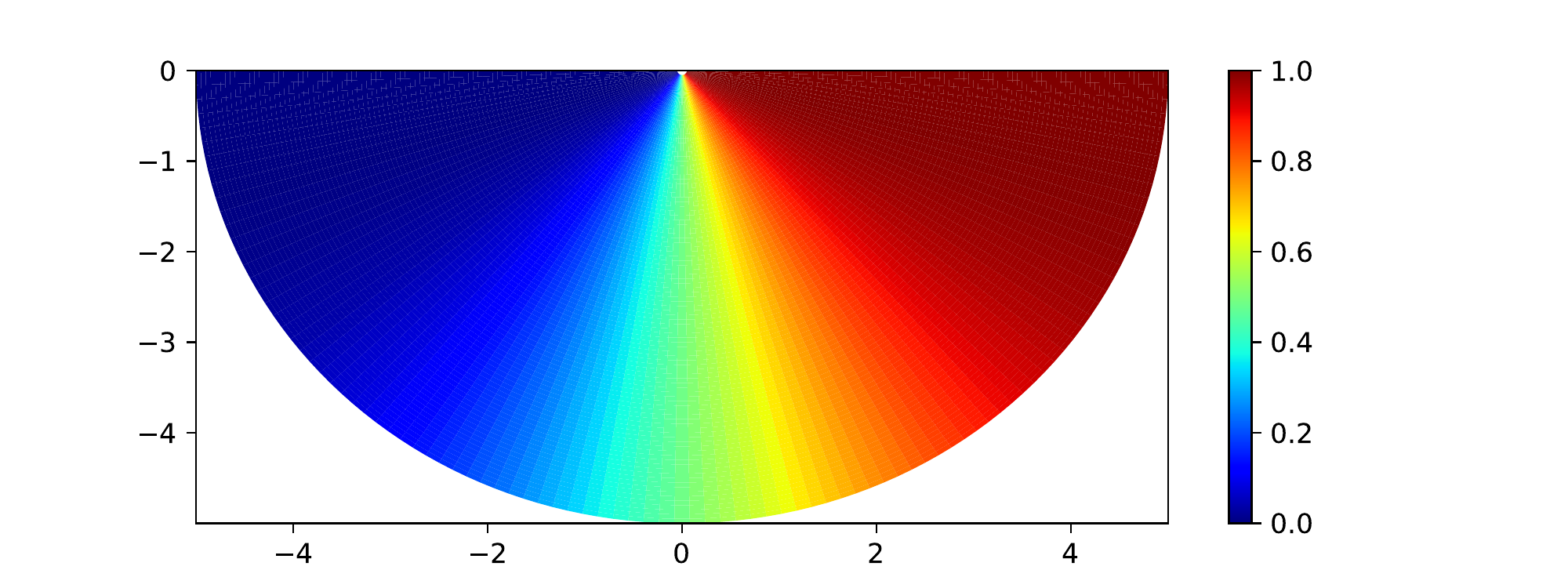}
    \caption{Reference solution}
    \label{fig:EX1-b}
\end{subfigure}
\caption{Contours of the vertical stresses $\sigma_y$ for half-space under semi-infinite uniform loading along one-half of its boundary  (color bar in MPa).}
\label{fig:Foundation_contours_1}
\end{figure}

\begin{figure}[!t]
\centering
\begin{subfigure}{0.4\textwidth}
    \centering
    \includegraphics[trim={0 0 0cm 0},width=0.99\linewidth]{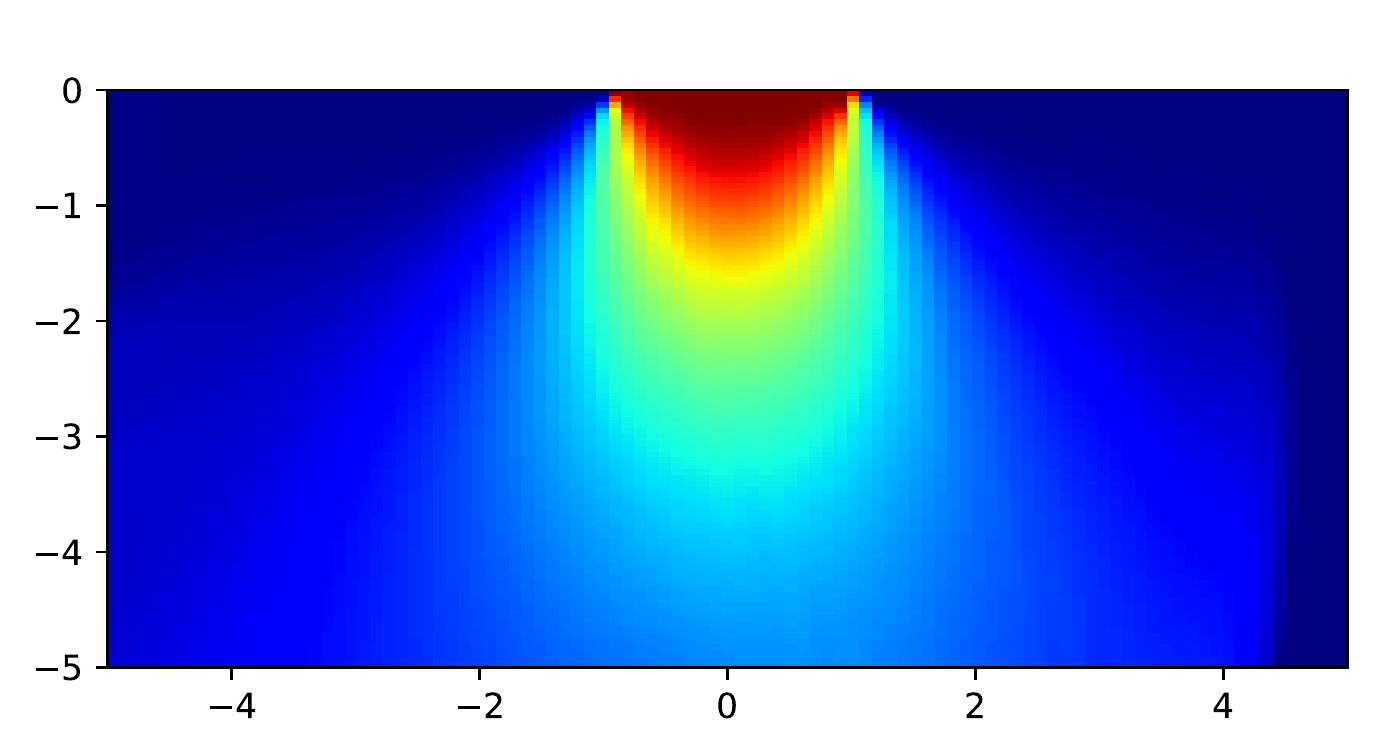}
    \caption{Conjugate networks solution}
    \label{fig:EX1-a}
\end{subfigure}
\begin{subfigure}{0.4\textwidth}
    \centering
    \includegraphics[trim={0 0 0cm 0},width=0.99\linewidth]{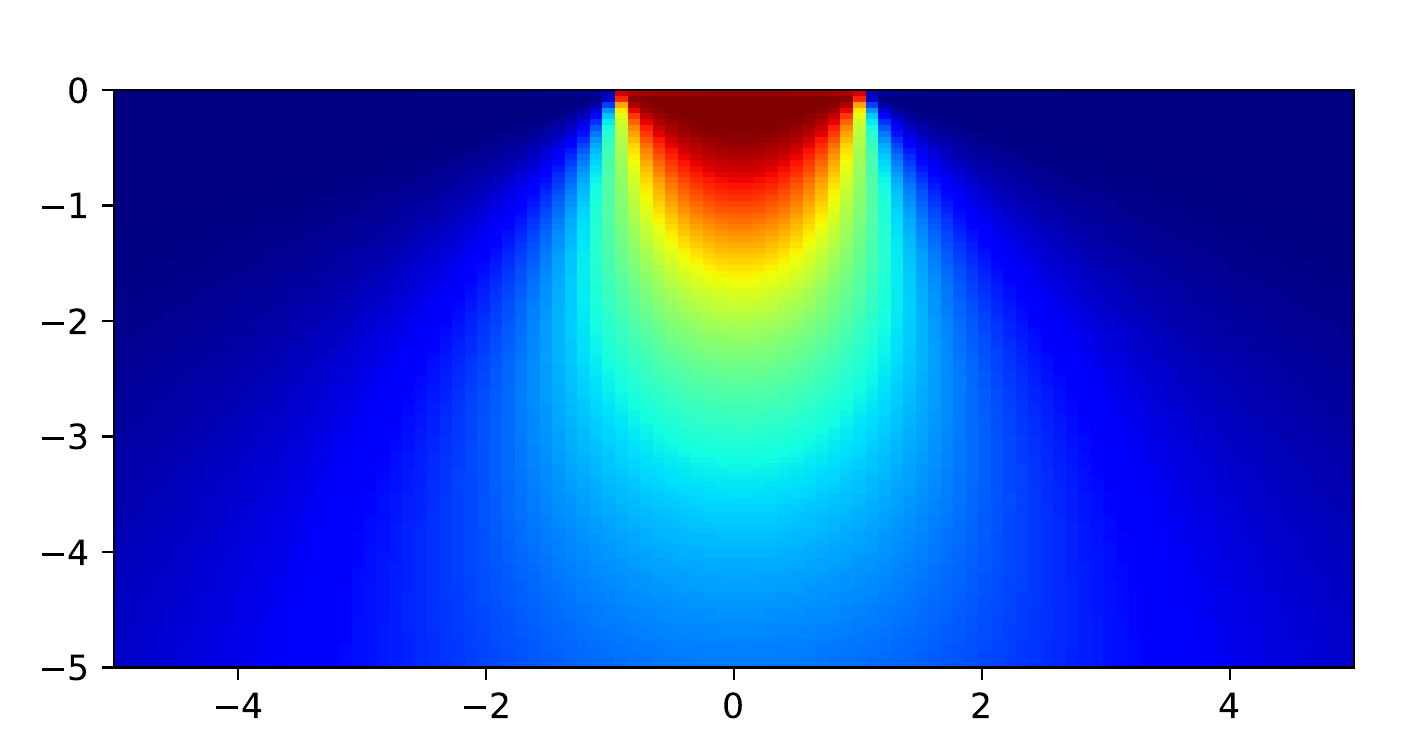}
    \caption{Airy-network solution}
    \label{fig:EX1-a}
\end{subfigure}
\begin{subfigure}{0.57\textwidth}
    \centering
    \includegraphics[trim={0 0 0cm 0},width=0.99\linewidth]{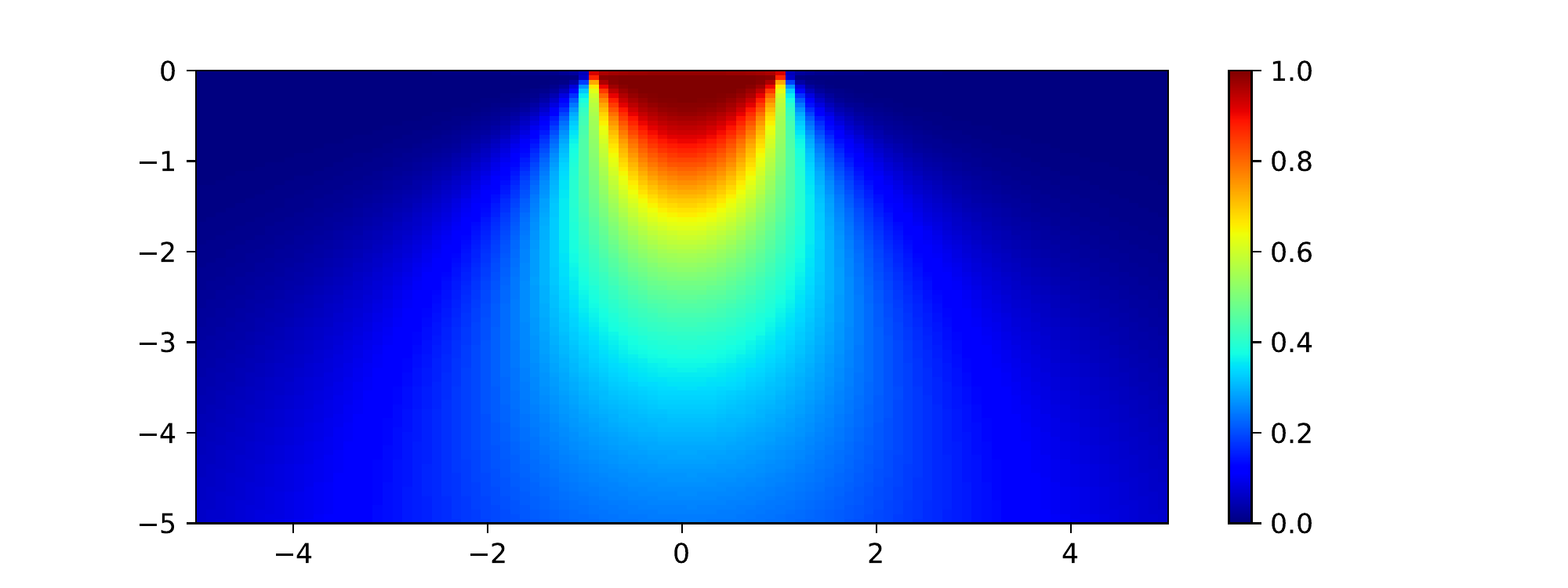}
    \caption{Reference solution}
    \label{fig:EX1-b}
\end{subfigure}
\caption{Contours of the vertical stresses $\sigma_y$ for the half-space subject to finite pressure (color bar in MPa).}
\label{fig:Foundation_contours_2}
\end{figure}

Stress contours for the half-space problem subjected to semi-infinite load are shown in Fig. \ref{fig:Foundation_contours_1} for both architectures and are compared to the analytical solution based on the theory of elasticity (see \cite{timoshenko412theory}). Note the stress components in polar coordinates are converted to Cartesian coordinates through stress transformation relations (e.g., see \cite{timoshenko1983history}). Evidently, the results corresponding to the Airy-network solution are in excellent agreement with the reference solution. On the contrary, slight discrepancies can be recognized in the results obtained by the conjugate neural networks. This is despite the beneficial use of the superposition rule which has substantially simplified the problem definition. Finally, in Fig. \ref{fig:Foundation_contours_2} stress contours for the half-space subjected to a finite interval of uniform pressure are presented for all solutions in conjunction with the reference solution. The results indicate the profound elaboration of Airy-network in the solution of the biharmonic equations in the theory of elasticity, which outperforms the more general physics-informed solutions.

 \section{Application of PINNs to theory of plates}
\label{S:4 (plate result)}
In this section, the theory of elastic thin plates is briefly reviewed. Moreover, the application of the PINNs approach in the solution to elastic plates is investigated in the framework of two benchmark examples.
\subsection{Theory of Elastic Plates}
\label{S:4-1 review}

\begin{figure}[!b]
\centering
\begin{minipage}{0.8\textwidth}
    \centering
    \includegraphics[width=0.7\linewidth]{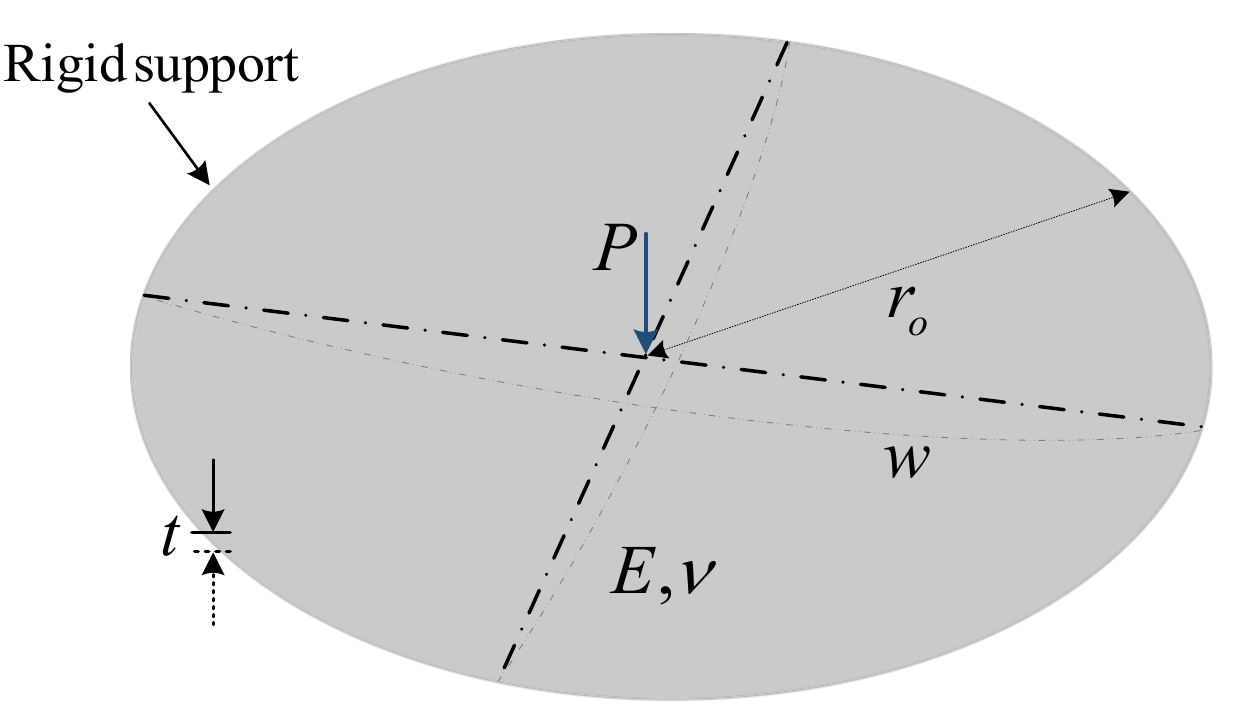}
    \end{minipage}
\caption{The problem definition and boundary conditions for circular plate under concentrated loading.}
    \label{fig:definition-circular}
\end{figure}

Plates are structural planar elements of relatively small thickness with respect to their dimensions. The plates theory takes advantage of the disparity in scales to simplify the three-dimensional plate problems into 2D ones. Here the classical Kirchhoff-Love's theory \cite{kirchhoff1850gleichgewicht,love1888xvi} for thin plates subjected to lateral loading is adopted, in which the transverse shearing strains are disregarded for the sake of simplicity. This renders the governing differential equations for elastic plates into a biharmonic partial differential equation (PDE), which is expressed in Cartesian coordinates as (see \cite{timoshenko1959theory})
\begin{equation} 
\begin{gathered}
\label{eq:platesCar}
\mathbf{\nabla} ^ {4} w = \left( {\frac{{{\partial ^2}}}{{\partial {x^2}}} +{\frac{\partial ^2}{\partial y^2}}} \right)\left( {\frac{{{\partial ^2}}}{{\partial {x^2}}} +{\frac{\partial ^2}{\partial y^2}}} \right)  w = \frac{q}{D}. \\
\end{gathered}
\end{equation}
In the above relation, $D$ denotes the flexural rigidity, $w$ is the lateral deflection, and $q$ is the distributed load over the plate. Note that the flexural rigidity $D$ is defined as $Et^3/[12(1-\nu^2)]$, in which $E$ is Young's modulus of elasticity, and $t$ is the plate thickness. In polar coordinates Eq. \eqref{eq:platesCar} transforms into (see \cite{reddy2006theory})
\begin{equation}
\label{eq:platesP}
\nabla ^4 w = \left( {\frac{{{\partial ^2}}}{{\partial {r^2}}} + \frac{1}{r}\frac{\partial }{{\partial r}} + \frac{1}{{{r^2}}}\frac{{{\partial ^2}}}{{\partial {\theta ^2}}}} \right)\left( {\frac{{{\partial ^2}}}{{\partial {r^2}}} + \frac{1}{r}\frac{\partial }{{\partial r}} + \frac{1}{{{r^2}}}\frac{{{\partial ^2}}}{{\partial {\theta ^2}}}} \right) w =\frac{q}{D}.
\end{equation}
The moments and shear forces are described versus plate deflection $w$ as 
\begin{equation}
\label{eq:moment-force}
\begin{split}
&{M_x = - D \ (\frac{\partial^ 2 w}{\partial x ^2} + \nu\frac{\partial^2 w} {\partial^ 2 y})}, \\
&{M_y = - D \ (\frac{\partial^ 2 w}{\partial y ^2} + \nu\frac{\partial^2 w} {\partial^ 2 x})}, \\
&{M_{xy} =  D \ (1-\nu)\frac{\partial^ 2 w}{\partial x \partial y}}, \\
&{Q_x = - D \ \frac{\partial}{\partial x}(\frac{\partial^ 2 w}{\partial x ^2} + \frac{\partial^2 w} {\partial^ 2 y})}, \\ 
&{Q_y = - D \ \frac{\partial}{\partial y}(\frac{\partial^ 2 w}{\partial x ^2} + \frac{\partial^2 w} {\partial^ 2 y})}, \\
\end{split}
\end{equation}
in which, $M_x$ and $M_y$ are bending moments, $M_{x y}$ is twisting moment, and $Q_x$ and $Q_y$ are shearing forces.

\subsection{Circular Plate Under Concentrated Loading}
\label{S:5-3 example-axisymetric-plate}
In this example, the PINNs solution of an axisymmetric circular plate subject to concentrated loading at its center is explored. The general form of the PDE governing the deflection of plates, given by Eq. \eqref{eq:platesP}, is reduced into a $3^\text{rd}$ order PDE for concentrated loads. Considering the axisymmetry of the problem, the solution becomes independent from $\theta$, and Eq. \eqref{eq:platesP} is further simplified into an ODE as \cite{timoshenko1959theory} 
\begin{equation}
\label{eq:ODE-polar-Qr}
{\frac{\partial^ 3 w}{\partial r ^3} + \frac{1}{r} \ \frac{\partial^2 w}{\partial r ^2} - \frac{1}{r^2} \ \frac{\partial w}{\partial r}= \frac{P}{2 \pi r D}},
\end{equation}
in which $P$ is the concentrated load.
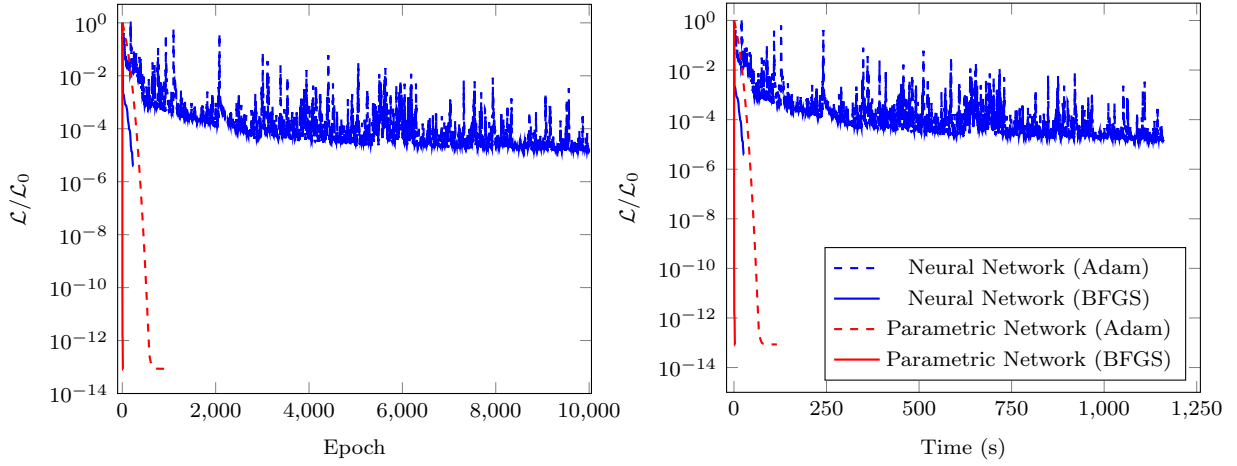
\begin{figure}[!t]
\centering
\begin{minipage}{0.48\textwidth}
    \begin{tikzpicture}
    \begin{axis}[
        every axis plot/.append      style={thick},
        ylabel={$\mathcal{L}/\mathcal{L_\text{0}}$},
        xlabel={Epoch},
        ymode=log,
        xmin=-100, xmax=10050,
        ymin=0.00000000000001, ymax=5.0,
        xtick={0,2000,4000,6000,8000,10000},
        ytick={0.00000000000001,0.000000000001,0.0000000001,0.00000001,0.000001,0.0001,0.01,1},
        legend pos=north east,
        /tikz/font=\footnotesize,
        legend style={font=\footnotesize},
        scaled x ticks = false,
        width = 0.99\linewidth,
        y tick label style={/pgf/number format/sci}
    ]
    \addplot[color=blue,dashed]
        table {./Data/loss_history_circularP_ANN.dat};
        \addplot[color=blue]
        table {./Data/loss_history_circularP_ANN_2nd.dat};
    \addplot[color=red,dashed]
        table {./Data/loss_history_circularP_Airy.dat};
        \addplot[color=red]
        table {./Data/loss_history_circularP_Airy_2nd.dat};
    \end{axis}
    \end{tikzpicture}
\end{minipage}
\begin{minipage}{0.48\textwidth}
    \begin{tikzpicture}
    \begin{axis}[
        every axis plot/.append      style={thick},
        ylabel={$\mathcal{L}/\mathcal{L_\text{0}}$},
        xlabel={Time (s)},
        ymode=log,
        xmin=-20, xmax=1260,
        ymin=0.000000000000001, ymax=5.0,
        xtick={0,250,500,750,1000,1250},
        ytick={0.00000000000001,0.000000000001,0.0000000001,0.00000001,0.000001,0.0001,0.01,1},
        legend pos=south east,
        /tikz/font=\footnotesize,
        legend style={font=\footnotesize},
        scaled x ticks = false,
        width = 0.99\linewidth,
        y tick label style={/pgf/number format/sci}
    ]
    \addplot[color=blue,dashed]
        table {./Data/loss_time_circularP_ANN.dat};
        \addlegendentry{Neural Network (Adam)};
    \addplot[color=blue]
        table {./Data/loss_time_circularP_ANN_2nd.dat};
        \addlegendentry{Neural Network (BFGS)};
    \addplot[color=red,dashed]
        table {./Data/loss_time_circularP_Airy.dat};
        \addlegendentry{Parametric Network (Adam)};
        \addplot[color=red]
        table {./Data/loss_time_circularP_Airy_2nd.dat};
        \addlegendentry{Parametric Network (BFGS)};
    \end{axis}
    \end{tikzpicture}
\end{minipage}
\caption{Training data history for circular plate under concentrated loading.}
\label{fig:loss-circular-plate}
\end{figure}

\begin{figure}[!b]
\centering
\begin{subfigure}{0.28\textwidth}
    \centering
    \includegraphics[trim={0 0 0cm 0},width=0.99\linewidth]{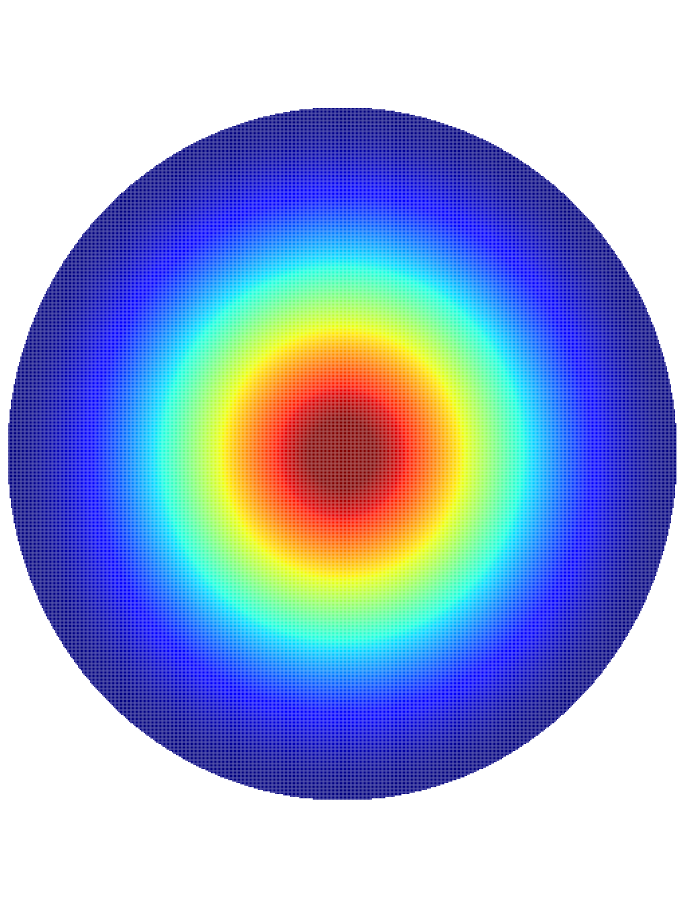}
    \caption{Neural network solution}
    \label{fig:EX1-a}
\end{subfigure}
\begin{subfigure}{0.29\textwidth}
    \centering
    \includegraphics[trim={0 0 0cm 0},width=0.99\linewidth]{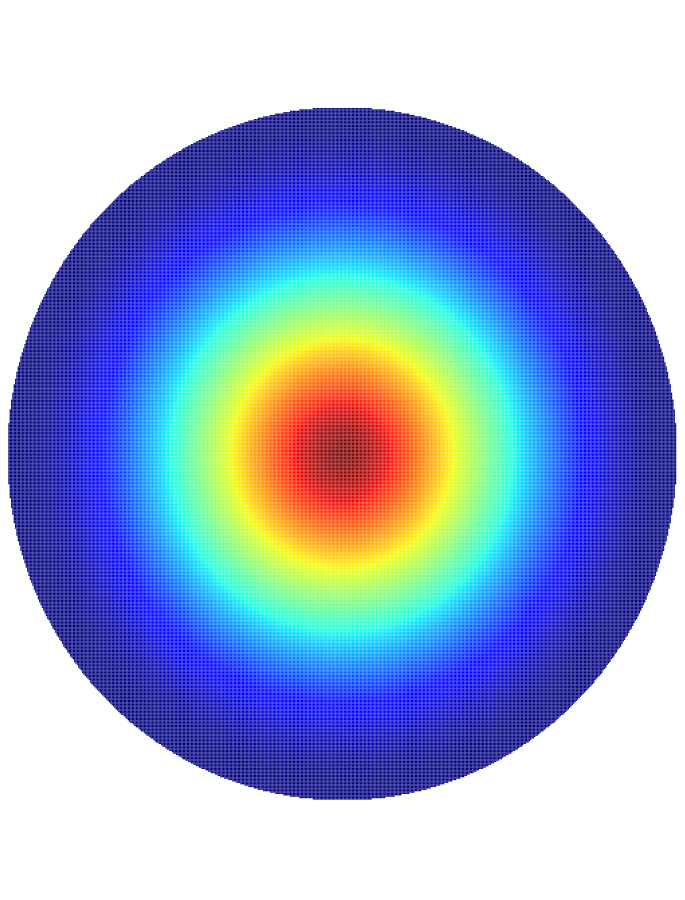}
    \caption{Parametric network solution}
    \label{fig:EX1-a}
\end{subfigure}
\begin{subfigure}{0.37\textwidth}
    \centering
    \includegraphics[trim={0 0 0cm 0},width=0.99\linewidth]{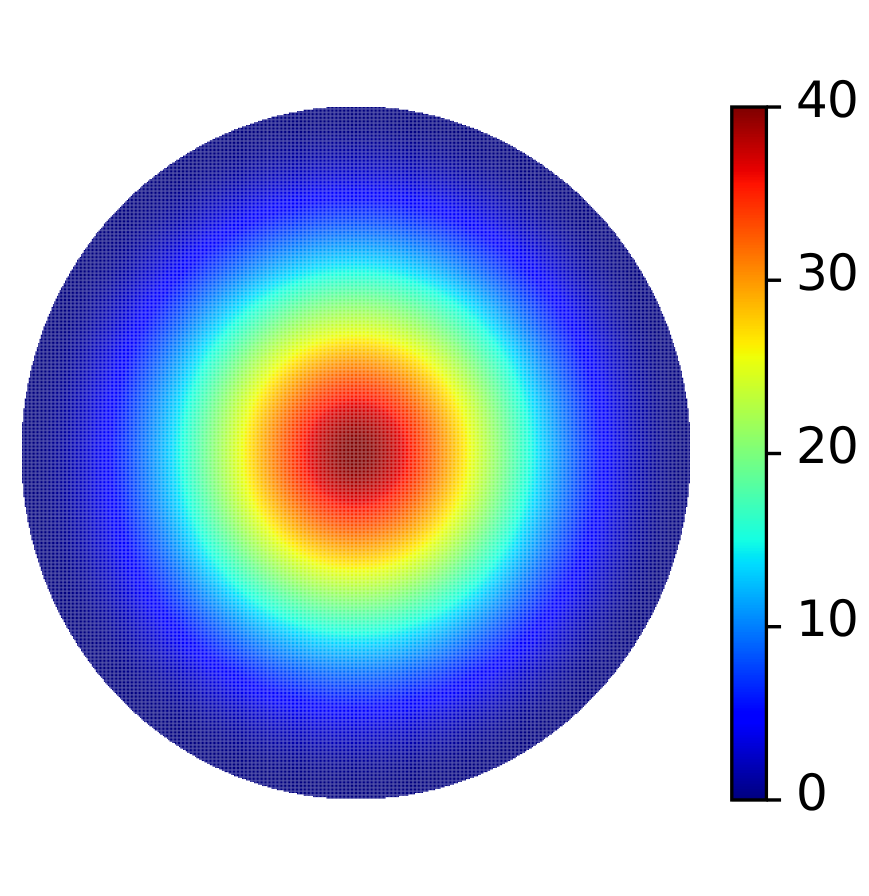}
    \caption{Reference solution $\,\,\,\,\,\,\,\,\,\,\,\,\,\,$}
    \label{fig:EX1-b}
\end{subfigure}
\begin{subfigure}{0.38\textwidth}
    \centering
    \includegraphics[trim={0 0 0cm 0},width=0.99\linewidth]{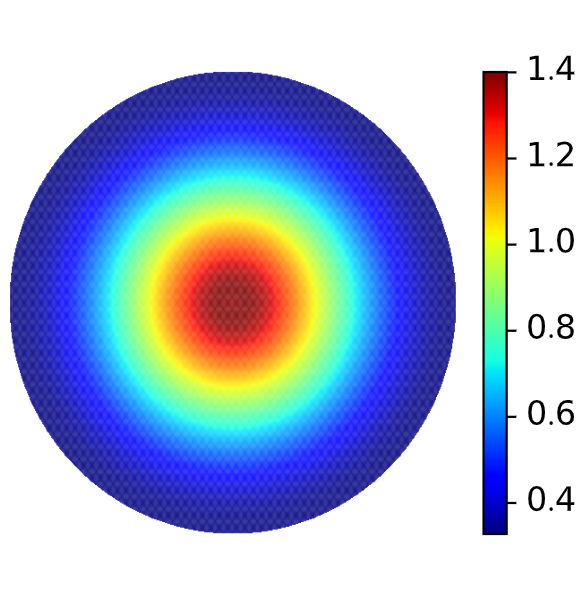}
    \caption{Absolute error of neural network}
    \label{fig:EX1-a}
\end{subfigure}
\begin{subfigure}{0.38\textwidth}
    \centering
    \includegraphics[trim={0 0 0cm 0},width=0.99\linewidth]{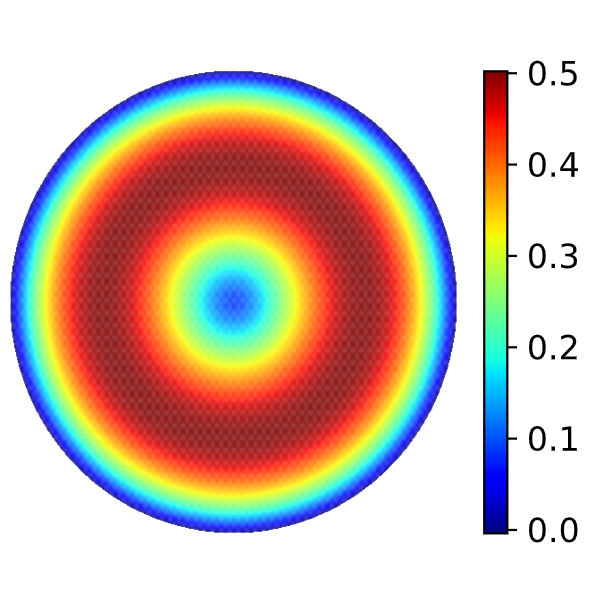}
    \caption{Absolute error of parametric network}
    \label{fig:EX1-a}
\end{subfigure}
\caption{Deflection contours for circular plate under concentrated loading in polar coordinates (color bar in mm).}
\label{fig:RESULT-CIRCULAR}
\end{figure}

As depicted in Fig. \ref{fig:definition-circular}, suppose a circular plate of radius $r_o= 1\ \text{m}$ and thickness $t = 0.03\ \text{m}$, subject to the concentrated load $P = 100 \ \text{kN} $, with rigid boundary condition at its periphery, i.e., $w(r=r_o,\theta)=\partial w / \partial r(r=r_o,\theta)= 0$. Considering the axisymmetry of the problem, the additional constraint of no angular deflection needs to be imposed to the problem at center, i.e., $\partial w/\partial r(r=0,\theta)=0$. The material properties of the plate are set to: Young's module of elasticity, $E = 20 \text{ GPa}$; and Poisson's ratio, $\nu = 0.25$.

The solution is constructed by definition of $w=\mathcal{N}{_w}(r)$, using 10 hidden layers with 10 neurons per layer. The network training is performed by using tanh as the activation function. Four loss functions are introduced to set up the optimization problem as
\begin{equation}
\begin{split}
\label{eq:loss_sum}
\mathcal{L}_\Omega&=\left\|  r^2\frac{\partial^ 3 w}{\partial r ^3} + r\frac{\partial^2 w}{\partial r ^2} - \frac{\partial w}{\partial r}- \frac{Pr}{2 \pi D} \right\|,\\
\mathcal{L}_{\partial\Omega_1}&=\lambda_1 \left\|(\partial w / \partial r)\left(r=0\right)\right\|,\\
\mathcal{L}_{\partial\Omega_2}&=\lambda_2 \left\|(\partial w / \partial r)\left(r=r_o\right)\right\|,\\
\mathcal{L}_{\partial\Omega_3}&=\lambda_3 \left\|( w)\left(r=r_o\right)\right\|.\\
\end{split}
\end{equation}
Note in the evaluation of domain losses $\mathcal{L}_\Omega$, to annihilate singularities with respect to $r$ at the origin, Eq. \eqref{eq:ODE-polar-Qr} is multiplied by $r^2$. As an alternative solution, for the case of uniform loading $q$, Eq. \eqref{eq:platesP} can be directly integrated into \cite{kelly2013solid}
\begin{equation}
\label{eq:PDE-plate-cte}
{w=-\frac{qr^4}{64D} + \frac{1}{4} a_1 r^2 \left( \ln r-1\right) + \frac{1}{4} a_2 r^2+a_3 \ln r+a_4},
\end{equation}
where for the concentrated loading case considered here the distributed load is vanished (i.e., $q=0$). Note $a_i$’s are the only parameters of the optimization problem identified exclusively by minimizing the loss functions related to the boundary conditions. The parametric network solution requires the introduction of an additional boundary condition related to shearing forces (i.e., $\mathcal{L}_{\partial\Omega_4}$) to accommodate the inclusion of the fourth unknown. Thus, the losses associated with the parametric solution are expressed as
\begin{equation}
\begin{split}
\label{eq:loss_sum2}
\mathcal{L}_{\partial\Omega_1}&=\lambda_1 \left\|(\partial w / \partial r)\left(r=0\right)\right\|,\\
\mathcal{L}_{\partial\Omega_2}&=\lambda_2 \left\|(\partial w / \partial r)\left(r=r_o\right)\right\|,\\
\mathcal{L}_{\partial\Omega_3}&=\lambda_3 \left\|( w)\left(r=r_o\right)\right\|,\\
\mathcal{L}_{\partial\Omega_4}&=\lambda_4 \| Q_r
-P/(2\pi\,r)\|,\\
\end{split}
\end{equation}
where,
\begin{equation}
\label{eq:Qr}
Q_r=D\left(\partial^3 w / \partial r^3+(1/r)\,\partial^2 w / \partial r^2-(1/r^2)\,\partial w / \partial r\right).
\end{equation}
Considering the singularity of Eqs. \eqref{eq:PDE-plate-cte} to \eqref{eq:Qr}  with respect to $r$ at origin, $r=0$ is excluded from the training of the parametric network.

Training is performed by means of a uniform grid consisting of 100 coalition points and the batch size of 32. Here, the Adam and BFGS optimisation schemes are elaborated for the training. The convergence history versus epochs and training time is plotted for the both networks in Fig. \ref{fig:loss-circular-plate}. Evidently, the BFGS scheme has outperformed the Adam algorithm in all cases. The convergence profile of the parametric network represents excellent convergence rate, that has reached at machine precision within relatively lesser epochs. In contrast, the relative error norm associated with the neural network solution lies in the range of $10^{-4}$ to $10^{-5}$, even beyond much more epochs (e.g., up to 10,000 epochs in the Adam scheme). This coincides with the longer training time required in the neural network solution.

Finally, in Fig. \ref{fig:RESULT-CIRCULAR} contours of the plate deflection are presented for both networks using the BFGS optimization scheme, in conjunction with the analytical solution for reference. In addition, the absolute error associated with each solution is plotted. Evidently, both networks demonstrate satisfactory accuracy. Still, the results related to the parametric network have been meaningfully more accurate with respect to the neural network solution. This further manifests the prominence of the alternative parametric approach in the solution of plate problems.

\subsection{Rectangular Plate Subject to Distributed Loading} 
\label{S:5-2 example-rectangular}
\begin{figure}[!t]
\centering
\begin{minipage}{0.8\textwidth}
    \centering
    \includegraphics[width=0.8\linewidth]{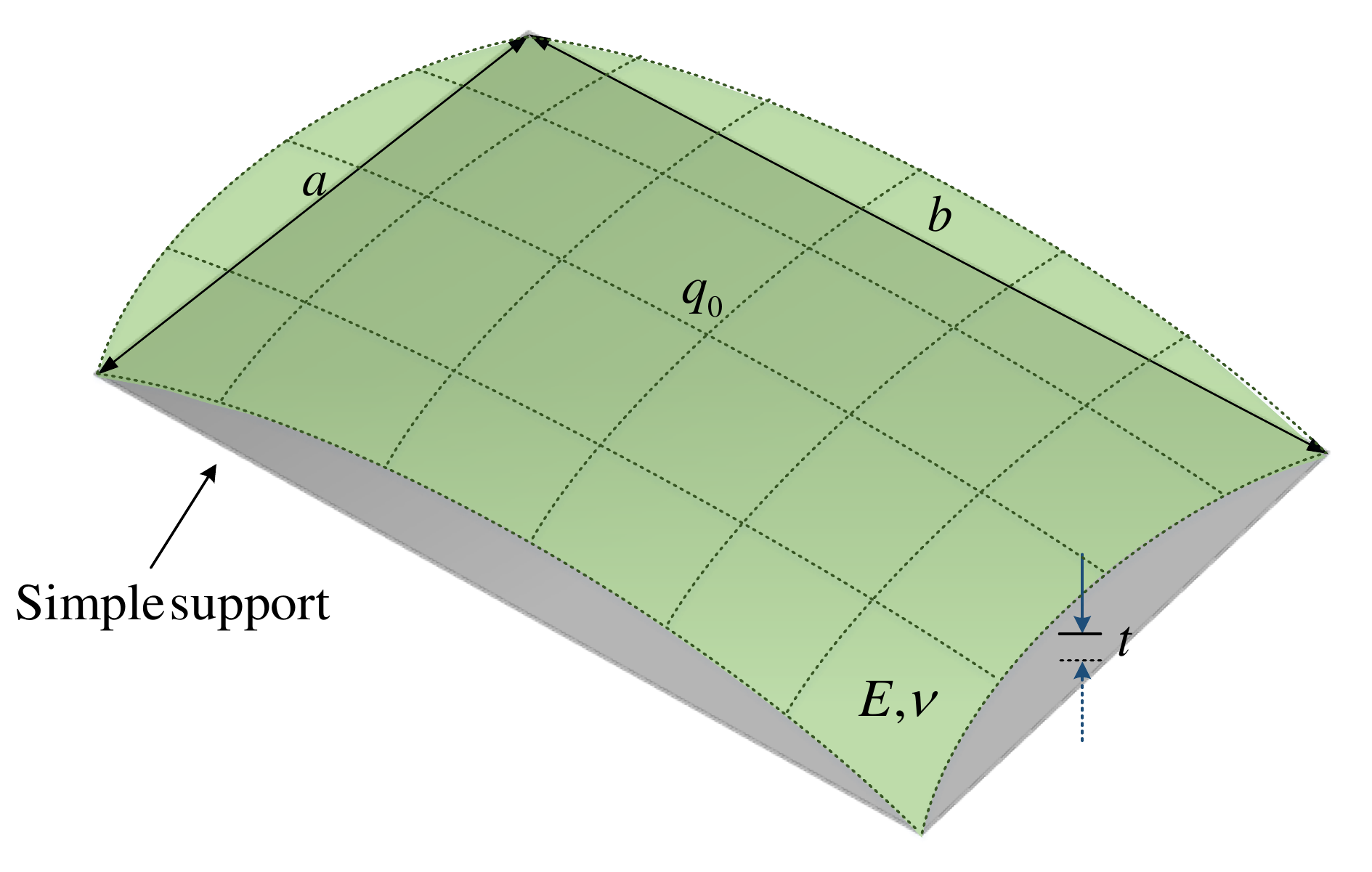}
    \end{minipage}
\caption{Problem definition and boundary conditions for simply supported rectangular plate subject to sinusoidal load.}
    \label{fig:definition-rectangular}
\end{figure}

In the final example, the application of the PINNs is demonstrated in the study of a simply supported rectangular plate under sinusoidal load (see Fig. \ref{fig:definition-rectangular}). The boundary conditions for the simple supports at the edges is described as
\begin{equation}
\begin{split}
\label{eq:q0-sin-load}
w(x=0, a; y)&=0,\\
w(x; y=0, b)&=0,\\
M_x(x; y=0, b)&=0,\\
M_y(x=0, a;y)&=0,
\end{split}
\end{equation}
where the distributed force exerted to the plate is expressed by
\begin{equation}
\label{eq:q0-sin-load}
{q= q_0 \sin(\frac{\pi x}{a}) \sin(\frac{\pi y}{b})}.
\end{equation}
The exact solution of plate deflection, bending and twisting moments, and shearing forces is summarized as \cite{timoshenko1959theory}:
\begin{equation}
\label{eq:EXACT_solution}
\begin{split}
w^* &=\frac{q_0}{\pi^4 D(\frac{1}{a^2}+\frac{1}{b^2})^2}\sin(\frac{\pi x}{a})\sin(\frac{\pi y}{b}), \\
M_x^* &=\frac{q_0}{\pi^2 (\frac{1}{a^2}+\frac{1}{b^2})^2}
\ (\frac{1}{a^2}+\frac{\nu}{b^2}) \sin(\frac{\pi x}{a}) \sin(\frac{\pi y}{b}), \\
M_y^* &=\frac{q_0}{\pi^2 (\frac{1}{a^2}+\frac{1}{b^2})^2}
\ (\frac{\nu}{a^2}+\frac{1}{b^2}) \sin(\frac{\pi x}{a}) \sin(\frac{\pi y}{b}), \\
M_{xy}^* &=\frac{q_0 \ (1-\nu)}{\pi^2(\frac{1}{a^2}+\frac{1}{b^2})^2 \ ab}
\cos(\frac{\pi x}{a}) \cos(\frac{\pi y}{b}), \\
Q_x^* &=\frac{q_0}{\pi a (\frac{1}{a^2}+\frac{1}{b^2})}
\cos(\frac{\pi x}{a}) \sin(\frac{\pi y}{b}), \\
Q_y^* &=\frac{q_0}{\pi b(\frac{1}{a^2}+\frac{1}{b^2})}
\sin(\frac{\pi x}{a}) \cos(\frac{\pi y}{b}). \\
\end{split}
\end{equation}
Here, a $200\text{ cm}\times  300 \text{ cm} $ rectangular plate with thickness $ t = 1 \text{ cm}$ is considered, and is subject to a sinusoidal load with intensity $q_0 = 0.01 \text{ kgf}/ \text{cm}^2$. The material properties of the plate are: Young's modulus of elasticity, $E = 2.06 \times 10^6 \text{ kgf}/ \text{cm}^2$; Poisson’s ratio, $\nu=0.25$; and flexural rigidity, $D=183111\text{ kgf.cm}$ .

The solution variable set involves deflection, moments and forces, in Cartesian coordinates, i.e., $w$, $M_x$, $M_y$, $M_{xy}$, $Q_x$, $Q_y$. These variables are defined as independent neural networks with the architecture of choice
\begin{equation}
\label{eq:approx_NN}
\begin{split}
w(x,y) &\simeq \mathcal{N}_{w}(x,y), \\
M_x(x,y) &\simeq \mathcal{N}_{M_x}(x,y), \\
M_y(x,y) &\simeq \mathcal{N}_{M_y}(x,y), \\
M_{xy}(x,y) &\simeq \mathcal{N}_{M_{xy}}(x,y), \\
Q_x(x,y) &\simeq \mathcal{N}_{Q_x}(x,y), \\
Q_y(x,y) &\simeq \mathcal{N}_{Q_y}(x,y).\\
\end{split}
\end{equation}
A network architecture of five layers with twenty neurons in each layer is used in this problem. The loss function is expressed as
\begin{equation}
\label{eq:loss_sum}
\begin{gathered}
\begin{split}
\mathcal{L}_{\Omega^*} &=\left\|w - w^*\right\| \\
                       &+\left\|M_x - M_x^*\right\|+\left\|M_y - M_y^*\right\|+\left\|M_{xy} - M_{xy}^*\right\| \\
                       &+\left\|Q_x - Q_x^*\right\|+\left\|Q_y - Q_y^*\right\|,\\
\mathcal{L}_{\Omega} &=\left\|\frac{\partial^ 4 w}{\partial x ^4} + 2 \ \frac{\partial^4 w}{\partial x ^2 \ \partial y ^2} + \frac{\partial^4 w}{\partial y^4} - \frac{q^*}{D}\right\|, \\
\mathcal{L}_{\partial\Omega} &= \left\| M_x + D \ (\frac{\partial^ 2 w}{\partial                                 x ^2} + \nu\frac{\partial^2 w} {\partial                                  y^2})\right\|\\
                            &+\left\|M_y + D \ (\frac{\partial^ 2 w}{\partial y ^2}+ \nu\frac{\partial^2 w} {\partial x^ 2})\right\|\\
                            &+\left\|M_{xy} - D \ (1-\nu)\frac{\partial^ 2 w}{\partial x \partial y}\right\|.
\end{split}
\end{gathered}
\end{equation}
In the above relations, the asterisk denotes the given data related to the exact solution (i.e., Eq. \eqref{eq:EXACT_solution}). In contrast, the quantities without asterisk express the approximation functions of the neural network (i.e., Eq. \eqref{eq:approx_NN}). Training is accomplished once the output values of all networks are as close as possible to the given data. The cost function given by Eq. \eqref{eq:loss_sum} can either be used for the solution of the plate deflection or identification of the model parameters. For the case of parameter identification here, $\nu$ and $D$ are treated as unknown network parameters that are attainable by using the deep-learning-based solution. This task can be easily carried out in SciANN with minimal coding (e.g., see \cite{haghighat2020deep}).

\begin{figure}[!b]  
\centering
\begin{minipage}{0.99\textwidth}
    \centering
    \includegraphics[width=0.99\linewidth]{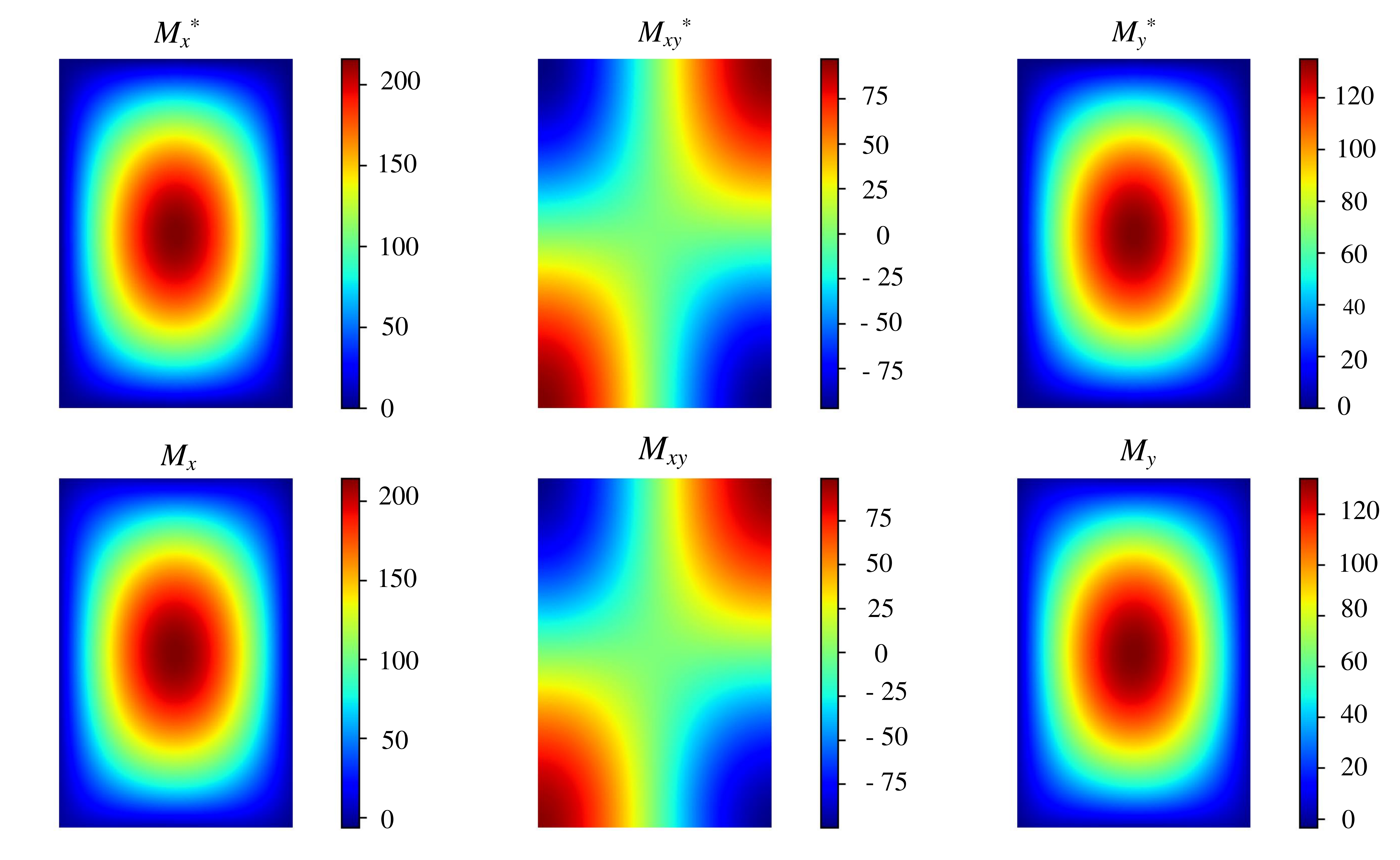}
    \end{minipage}
\caption{A comparison between the PINNs solution and exact values of bending and twisting moment of rectangular plate subject to distributed loading (color bar in SI units).}
    \label{fig:moment*AND-PNN}
\end{figure}

\begin{figure}[!b]  
\centering
\begin{minipage}{0.99\textwidth}
    \centering
    \includegraphics[width=0.99\linewidth]{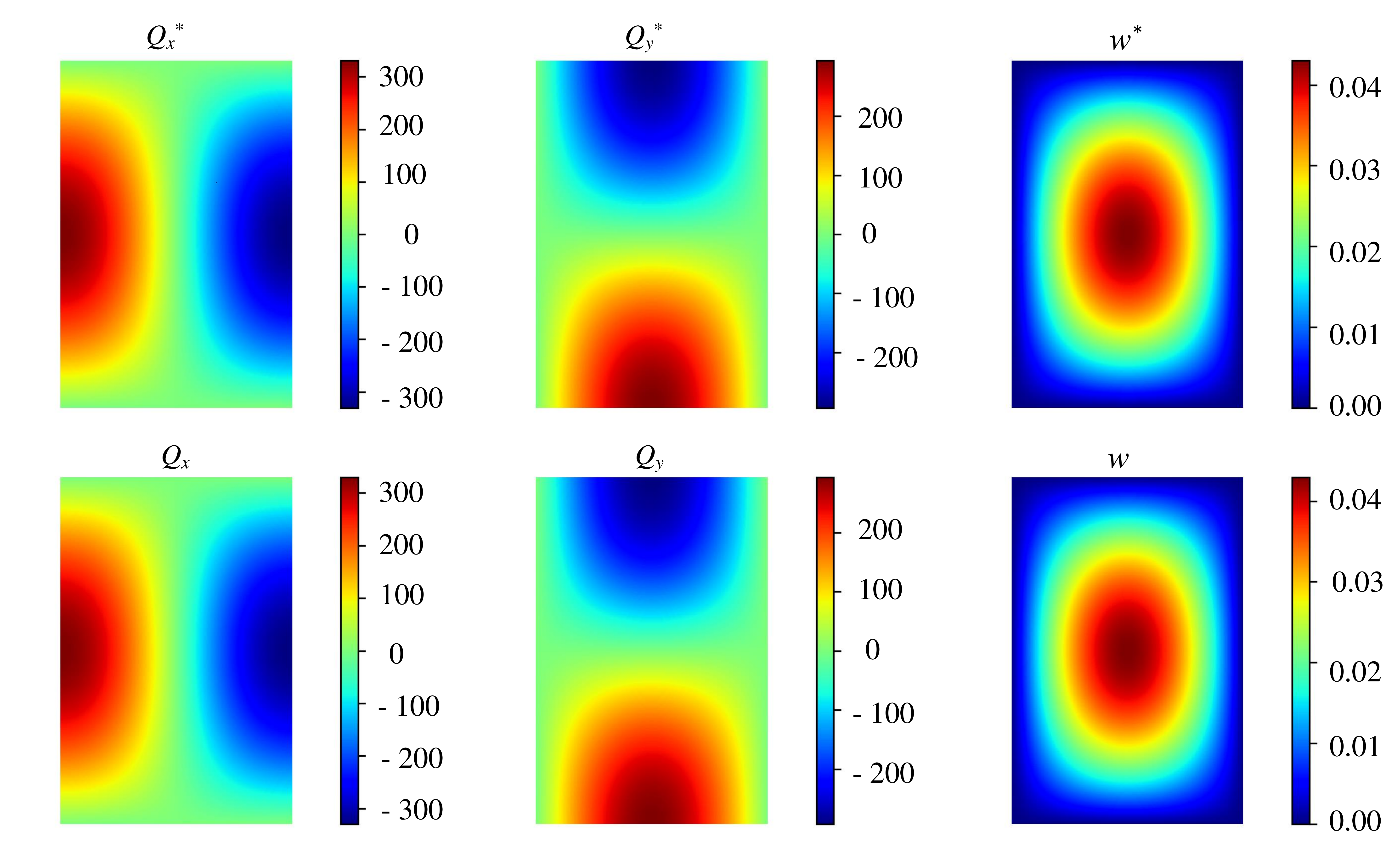}
    \end{minipage}
\caption{A comparison between the PINNs solution and exact values of shearing forces and deflection of rectangular plate subject to distributed loading (color bar in SI units).}
    \label{fig:FORCE*-PNN}
\end{figure}

The dataset consists of a $30 \times 30$ uniformly distributed sample points. The learning rate and batch size of the simulation are 0.001 and 50, respectively, where the activation function tanh is utilized. The results of plate deflection, moments, and shear forces obtained from the neural network approximation as well as the exact analytical solutions are shown in Fig. \ref{fig:moment*AND-PNN} and Fig. \ref{fig:FORCE*-PNN}. Evidently, an excellent agreement can be seen between both solutions.

Alternatively, the identification of the model parameters is carried out through the elaboration of the exact solutions of the plate deflection given by Eq. \eqref{eq:EXACT_solution}. Similar sampling points are applied for the training of the neural network. Evaluated parameters (i.e., the Poisson's ratio and flexural rigidity) obtained from the PINNs solution are presented in Table \ref{table: discovery-rectangular0}, and compared to the exact values. Evidently, the errors are extremely negligible for both parameters (i.e., less than 0.25\%). Notably, the role of the available data on this process is crucial since parameter identification is in particular of interest with as little data as possible. Using the PINNs solution here, the model parameters are impeccably determined by the use of merely $900$ datasets. 

\begin{figure}[!ht]
\begin{minipage}{0.49\textwidth}
    \centering
    \begin{tikzpicture}
    \begin{axis}[
        every axis plot/.append style={thick},
        ylabel={$\mathcal{L}/\mathcal{L_\text{0}}$},
        xlabel={Epochs},
        ymode=log,
        xmin=-20, xmax=1220,
        ymin=10^-5, ymax=10.0,
        xtick={0,2*10^2,400,600,800,1000,1200},
        ytick={1e-5,1e-4,1e-3,1e-2,1e-1,1},
        legend pos=north east,
        /tikz/font=\footnotesize,
        legend style={font=\footnotesize},
        width = 0.99\linewidth
    ]
    \addplot[color=blue]
        table {./Data/loss_rectangular_Edit.dat};
    \addplot[color=red]
        table {./Data/loss-rectang-parameter.txt};
    \end{axis}
    \end{tikzpicture}
    \label{fig:LOSS-EPOCH}
\end{minipage}
\begin{minipage}{0.49\textwidth}
    \centering
    \begin{tikzpicture}
    \begin{axis}[
        every axis plot/.append      style={thick},
        ylabel={$\mathcal{L}/\mathcal{L_\text{0}}$},
        xlabel={Time(s)},
        ymode=log,
        xmin=-10, xmax=310,
        ymin=10^-5, ymax=10.0,
        xtick={0,60,120,180,240,300},
        ytick={1e-5,1e-4,1e-3,1e-2,1e-1,1},
        legend pos=north east,
        /tikz/font=\footnotesize,
        legend style={font=\footnotesize},
        width = 0.99\linewidth
    ]
    \addplot[color=blue]
        table {./Data/time_rectangular_Edit.dat};
        \addlegendentry{Solution study};
    \addplot[color=red]
        table {./Data/time-history-parameter.txt};
        \addlegendentry{Identification study};
    \end{axis}
    \end{tikzpicture}
    \label{fig:LOSS-TIME}
\end{minipage}
\caption{Training data history in the simply supported rectangular plate under sinusoidal load; a) convergence history of the cost function, b) network training time history.}
\label{fig:training_history_rectagular}
\end{figure}
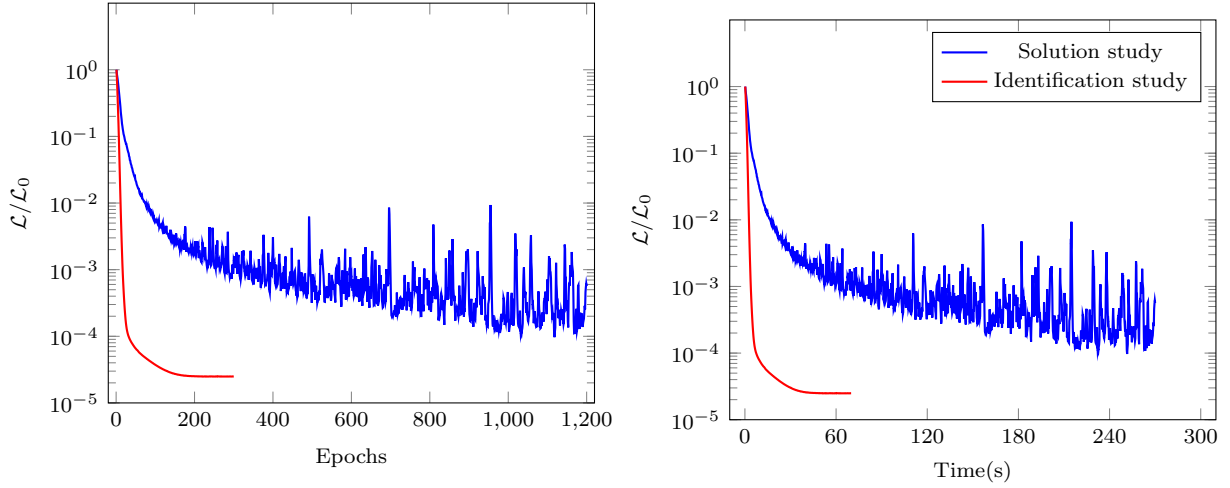

In Fig. \ref{fig:training_history_rectagular}, the convergence profile and time history of the cost function associated with training are plotted for both the solution and parameter identification studies. As can be seen, the parameter identification study has been relatively faster, whereas within less than 200 epochs and 1 minute of training the loss function has reached the relative error of $10^{-5}$. In contrast, the solution study has been more time-consuming with the relative error of $10^{-4}$ reached beyond 1200 epochs and 4 minutes of training. Still, both demonstrate promising accuracy in terms of predicted results.  

\begin{table}[!b]
\footnotesize
\centering
\caption{Identification study of rectangular plate parameters, $\nu$ and $D$.}
\begin{tabular}{ccc}
\Xhline{2\arrayrulewidth}
  {Parameters} & Analytical Solution & PINNs Solution \\ 
\Xhline{2\arrayrulewidth}
 {Poisson's ratio} & 0.2500 & 0.2497\\
 \hline
 {Flexural Rigidity} & 183111 & 183382 \\
\Xhline{2\arrayrulewidth}
\end{tabular}
\label{table: discovery-rectangular0}
\end{table}

At the end of this example, the PINNs is employed for the parametric study in rectangular plates. For this sake, the general loading condition, i.e., $q=f(x,y)$, is investigated which is known as Navier's \cite{Navier} problem. Accordingly, the double Fourier sine series is supposed to approximate the distributed load function $f(x,y)$ for which the explicit solution is available in resemblance to Eq. \eqref{eq:EXACT_solution}. For the case of uniform load distribution over the entire surface of the rectangular plate (i.e., $f(x,y)=q_0$), the closed-form solution is described as \cite{timoshenko1959theory}
\begin{equation}
\label{eq:deflection_SUM_q0}
w =\frac{16 \ q_0}{\pi^6 \ D}\ \sum_{m=1}^{k} \ \sum_{n=1}^{k}\
\frac{\sin\frac{m\pi x}{a} \sin\frac{n\pi y}{b}}{mn\ (\frac{m^2}{a^2} +\ \frac{n^2}{b^2})^2},\\
\end{equation}
in which $m$ and $n$ are odd integers. Note all terms related to either case where $m$ or $n$ is an even number vanish. It is noteworthy that the Fourier series lose their significance as $m$ and/or $n$ increase. Indeed, the expansion related to $n, m$ = 1, 3, 5 accommodates a near-exact solution accuracy (i.e., $99.9\% $ precision \cite{timoshenko1959theory}) to the plate deflection subject to uniform distributed loading $q_0$, which follows
\begin{equation}
\label{eq:deflection_SUM_q0}
\begin{gathered}
\begin{split}
w = \frac{16 \ q_0}{\pi^6 \ D}\ ( \frac{\sin\frac{\pi x}{a} \sin\frac{\pi y}{b}}{(\frac{1}{a^2}+\ \frac{1}{b^2})^2} &+ \frac{\sin\frac{\pi x}{a} \sin\frac{3\pi\ y}{b}}{3\ (\frac{1}{a^2} +\ \frac{9}{b^2})^2} +\
\frac{\sin\frac{\pi x}{a} \sin\frac{5\pi y}{b}}{5\ (\frac{1}{a^2} +\ \frac{25}{b^2})^2} +\\
\frac{\sin\frac{3\pi x}{a} \sin\frac{\pi y}{b}}{3(\frac{9}{a^2} +\ \frac{1}{b^2})^2} &+\
\frac{\sin\frac{3\pi x}{a} \sin\frac{3\pi y}{b}}{9(\frac{9}{a^2} +\ \frac{9}{b^2})^2}+\
\frac{\sin\frac{3\pi x}{a} \sin\frac{5\pi y}{b}}{15(\frac{9}{a^2} +\ \frac{25}{b^2})^2}+\\
\frac{\sin\frac{5\pi x}{a} \sin\frac{\pi y}{b}}{5(\frac{25}{a^2} +\ \frac{1}{b^2})^2} &+\
\frac{\sin\frac{5\pi x}{a} \sin\frac{3\pi y}{b}}{15(\frac{25}{a^2} +\ \frac{9}{b^2})^2}+\
\frac{\sin\frac{5\pi x}{a} \sin\frac{5\pi y}{b}}{25(\frac{25}{a^2} +\ \frac{25}{b^2})^2}).\\
\end{split}
\end{gathered}
\end{equation}
In order to conduct the parametric study by means of PINNs, this solution is recast into the following format 
\begin{equation}
\begin{gathered}
\begin{split}
\label{eq:deflection_SUM2_q0}
w = &a_1\ (\sin\frac{\pi x}{a} \sin\frac{\pi y}{b}) + a_2\ (\sin\frac{\pi    x}{a} \sin\frac{3\pi y}{b})+\
  a_3\ (\sin\frac{\pi x}{a} \sin\frac{5\pi y}{b})+\\
  &a_4\ (\sin\frac{3\pi x}{a} \sin\frac{\pi y}{b}) +a_5\ (\sin\frac{3\pi x}{a} \sin\frac{3\pi y}{b})+\
  a_6\ (\sin\frac{3\pi x}{a} \sin\frac{5\pi y}{b})+\\ 
  &a_7\ (\sin\frac{5\pi x}{a} \sin\frac{\pi y}{b})+\
  a_8\ (\sin\frac{5\pi x}{a} \sin\frac{3\pi y}{b})+\
  a_9\ (\sin\frac{5\pi x}{a} \sin\frac{5\pi y}{b}),\\
\end{split}
\end{gathered}
\end{equation}
in which $a_i$'s are model parameters to be determined upon the imposition of the BCs. In Table \ref{table: discovery-rectangular}, the parameters obtained via PINNs solution are presented and compared to the analytical values given by Eq. \eqref{eq:deflection_SUM_q0}. Evidently, an excellent agreement can be seen between both solutions, with relative errors in the most significant terms (i.e., the first four terms where $a_i>0.01a_1$) lies below 0.5\%. This further highlights the outstanding performance of the PINNs in parameter identification.

\begin{table}[!ht]
\footnotesize
\centering
\caption{Parametric solution of rectangular plate subject to uniform loading, $q_0$.}
\begin{tabular}{ccc}
\Xhline{2\arrayrulewidth}
  {Parameters} & Analytical Solution & PINNs Solution \\ 
\Xhline{2\arrayrulewidth}
 {$a_1$} & 6.969e-2 & 6.944e-2\\
 \hline
 {$a_2$} &1.939e-3 &1.927e-3 \\
 \hline
 {$a_3$} &1.981e-4  & 1.989e-4 \\
 \hline
 {$a_4$} &5.430e-4  & 5.433e-4 \\
 \hline
 {$a_5$} &9.56e-5 & 9.79e-5 \\
 \hline
 {$a_6$} &2.39e-5  & 2.31e-5 \\
 \hline
 {$a_7$} &4.49e-5  & 4.55e-5 \\
 \hline
 {$a_8$} &1.15e-5 & 1.22e-5 \\
 \hline
 {$a_9$} & 4.46e-6 & 3.57e-6 \\ 
\Xhline{2\arrayrulewidth}
\end{tabular}
\label{table: discovery-rectangular}
\end{table}


\section{Conclusions}
\label{S:5 (Conclusions)}

In this study, the application of Physics-Informed Neural Networks (PINNs) for the solution and parameter identification in the theory of elasticity and elastic plates is explored. Fundamentals of Physics-Informed deep learning is briefly explained in conjunction with the construction and training of neural networks. Thereafter, the application of PINNs to the theory of elasticity is studied. The Airy stress function and the biharmonic PDEs governing the elasticity solutions are presented in the general form. The application of PINNs in the solution of the Lamé Problem is demonstrated, as a representative benchmark example involving a simplified biharmonic ODE formulation in polar coordinates. In addition, an Airy-inspired parametric solution to the same problem is investigated, which is developed by a manufactured combination of a set of nonlinear terms based on the theory of elasticity. It is shown that Airy-network results in superior convergence and accuracy characteristics. The next example is dedicated to the foundation problem, in which the PINNs solution of a simplified PDE in polar coordinates is developed in conjunction with mapping in Cartesian coordinates. It is shown the alternative use of Airy-inspired neural networks outperforms the classic PINNs solution by a fair margin. The application of PINNs for solution and discovery in the theory of elastic plates is implemented next. The fourth-order PDE governing the lateral deflection of thin plates is explained concisely. A circular plate under concentrated loading is studied, which involves the application of PINNs to the solution of reduced third order ODE of plate deflection in polar coordinates. The manufactured solution inspired by the analytical approach is exercised and proven to be a promising alternative to the general PINN approach again. The final example is devoted to a rectangular plate subject to distributed loading. This example consists of both the solution and parameter identification of the generic fourth-order PDE of plate deflection. As a remedy, Fourier series are elaborated to investigate the solution of plates deflection, which demonstrates enormous improvement in terms of training duration and accuracy. In this fashion, it is shown that the classical analytical methods could contribute significantly to the construction of neural networks with a minimal number of parameters that are very accurate and fast to evaluate. Thus, they can be used to guide the construction of more efficient physics-informed neural networks. Considering that \emph{Fourier features} are now commonly employed to construct more trainable neural networks, a natural extension would be to also add \emph{Airy features} to improve the trainability of neural networks for problems of continuum mechanics.

\section*{Data Availability Statement}
All data, models, or source codes that support the findings of this study are available from the corresponding author upon reasonable request.

\bibliographystyle{model1-num-names.bst}
\bibliography{References.bib}

\end{document}